%% file: main.tex
\documentclass[11pt]{article}

\usepackage{longtable}
\usepackage{amssymb}
\usepackage{amsmath}
\newcommand{\prob}{\mathbb{P}}
\usepackage{adjustbox}
\usepackage[final]{acl}
\usepackage{subcaption}

\usepackage{times}
\usepackage{latexsym}

\usepackage[T1]{fontenc}
\usepackage{hyperref}
\usepackage[utf8]{inputenc}
\usepackage{microtype}

\usepackage{graphicx}

\usepackage{booktabs}
\usepackage{multirow}
\usepackage{colortbl}

\usepackage{txfonts}

\usepackage[status=final,nomargin,inline,lang=spanish]{fixme}
\FXRegisterAuthor{dg}{adg}{}

\newcommand{\careqa}{\texttt{CareQA}} 
\newcommand{\relaxed}{\texttt{Relaxed Perplexity}}

\title{Automatic Evaluation of Healthcare LLMs Beyond Question-Answering}

\author{
    \textbf{Anna Arias-Duart\textsuperscript{\dag}\textsuperscript{1}},
    \textbf{Pablo Agustin Martin-Torres\textsuperscript{\dag}\textsuperscript{1}},
    \textbf{Daniel Hinjos\textsuperscript{1}},
\\
    \textbf{Pablo Bernabeu-Perez\textsuperscript{1}},
    \textbf{Lucia Urcelay Ganzabal\textsuperscript{3}},
    \textbf{Marta Gonzalez Mallo\textsuperscript{1}},
\\
    \textbf{Ashwin Kumar Gururajan\textsuperscript{1}},
    \textbf{Enrique Lopez-Cuena\textsuperscript{1}},
\\
    \textbf{Sergio Alvarez-Napagao\textsuperscript{1,2}},
    \textbf{Dario Garcia-Gasulla\textsuperscript{1}}
\\
\\
    \textsuperscript{\dag} Equal contribution. \textsuperscript{1} Barcelona Supercomputing Center (BSC) \\
    \textsuperscript{2} Universitat Politècnica de Catalunya (UPC)–BarcelonaTech \\
    \textsuperscript{3} Independent Researcher (formerly affiliated with BSC) \\
}

\newcommand{\eg}{\emph{e.g.}, }       
\newcommand{\ie}{\emph{i.e.}, }      

\begin{document}
\maketitle
\begin{abstract}

Current Large Language Models (LLMs) benchmarks are often based on open-ended or close-ended QA evaluations, avoiding the requirement of human labor. Close-ended measurements evaluate the factuality of responses but lack expressiveness. Open-ended capture the model's capacity to produce discourse responses but are harder to assess for correctness. 
These two approaches are commonly used, either independently or together, though their relationship remains poorly understood. This work is focused on the healthcare domain, where both factuality and discourse matter greatly. It introduces a comprehensive, multi-axis suite for healthcare LLM evaluation, exploring correlations between open and close benchmarks and metrics. Findings include blind spots and overlaps in current methodologies. As an updated sanity check, we release a new medical benchmark ~---\careqa{}---~, with both open and closed variants. Finally, we propose a novel metric for open-ended evaluations ~---\relaxed{}---~ to mitigate the identified limitations.

\end{abstract}

\input{latex/tables/table_5}

\section{Introduction}


The growing use of large language models (LLMs) in public domains, such as healthcare, shows promise for improving global quality of life~\cite{he2023survey}. At the same time, the reliability and evaluation of LLMs in such sensitive topics requires extreme caution due to the potential impact on people's rights and well-being. 

LLM evaluation today is approached through various perspectives, which consider different types of LLM assessment: automatic evaluation (scalable and factual), user evaluation (utility and usability)~\cite{chiang2024chatbot}, and expert evaluation (support and coherence)~\cite{chen2023large}. While each of these evaluation perspectives serves distinct roles that contribute to a holistic assessment, automatic evaluation remains the most prevalent one due to its lack of dependency on human effort. 

Within automatic evaluation, there are two types of tests. Those which include closed-ended responses \cite{bedi2024systematic}, namely multiple-choice question answering (MCQA), and those which have open-ended responses \cite{dada2024clue}. Close-ended MCQA validation enables the automatic verification of response factuality, but it does not reflect the complex nature of real world situations (\eg clinical settings \cite{hager2024evaluation, zhou2023survey}). As such, MCQA alone often fails to identify critical short-comings of model performance~\cite{li2024can,umapathi2023med,ahmad2023creating,pezeshkpour2023large, alzahrani2024benchmarks, zheng2023large}. 

To incorporate a broader range of tasks relevant to the medical field \cite{dada2024clue, kanithi2024medic}, one typically has to rely on open-ended answers. That is, reference responses are not the only valid outputs. Since these cannot be completely assessed for factuality without human expert supervision, approximate measures based on n-grams and model perplexity remain in place, which limits the reliability of these  evaluations \cite{kamalloo2023evaluating}.

Efforts have been dedicated to analyze the relation between automatic evaluations and either user or expert evaluations, showing a lack of direct correspondence~\cite{fleming2024medalign,nimah2023nlg}. This is explained by the difference in the model features these assess (\eg factuality vs usability vs support capacity), pointing at their complementary nature. Nonetheless, a similar analysis within the family of automatic evaluations is still pending; a study of the relations between open-ended and close-ended benchmarks and metrics, to understand which of these tests should be used, and when. For that purpose, we focus on the healthcare domain, providing the following contributions:

\begin{itemize}
    \item A correlation-based, empirical analysis of open-ended and close-ended tasks, benchmarks, and metrics.
    \item A novel medical benchmark (\careqa{}) featuring both closed- and open-ended formats for the verification of our findings.
    \item A new metric for open-ended evaluations (\relaxed{}) which fills a gap identified in existing methodologies.
\end{itemize}

\section{Methodology}\label{sec:methodology}

This study considers four different close-ended healthcare tasks, which include nine different datasets (\eg MedQA). These are all assessed using the accuracy metric. At the same time, six open-ended tasks are studied, based on nine distinct datasets (\eg MedText). In this case, eleven different metrics are extracted. Further details are shown in Table~\ref{tab:tasks_bench_metrics}. To assess the consistency within tasks, datasets and metrics, this work considers up to 12 different open LLMs, both specifically tuned for healthcare and general purpose, motivated by previous work~\cite{shoham2024medconceptsqa, kanithi2024medic}.



\subsection{\careqa{}: A Novel Benchmark}\label{sec:careqa}



Updated benchmarks are necessary to prevent both data drift (as human knowledge evolves), and data contamination (as training data crawling efforts scale). To validate the integrity and consistency of existing tests, this work introduces a new benchmark for automatic evaluation, \careqa{}, available in both closed-ended and open-ended formats.

\careqa{} originates from the Spanish Specialised Healthcare Training (MIR) exams by the Spanish \textit{Ministry of Health}. The close-ended version is a MCQA including 5,621 QA pairs across six categories: medicine, nursing, biology, chemistry, psychology, and pharmacology, sourced from the 2020 to 2024 exam editions. \careqa{} is available in both English and Spanish, with the translation performed using GPT-4.

The open-ended version (English only) was created by rephrasing the questions from the close-ended version using the \href{https://huggingface.co/Qwen/Qwen2.5-72B-Instruct}{Qwen2.5-72B-Instruct} model. After the rephrasing process, the number of suitable questions was reduced to 3,730 QA pairs. This set retains the same categories as the closed-ended version. 

To ensure the validity of both the translations and rephrasing, 10 annotators conducted a manual review of a total of 360 samples, each reviewed by at least three evaluators. This process achieved a confidence level of 95\% and a margin of error of 5\% approximately.

The translation results were positive, with all three evaluators agreeing on 83.1\% of the questions as correct. Based on this, we considered the translation to be of good quality. However, the percentage of rephrased QA pairs labeled as correct by the three evaluators was 65.8\%. 

To address this, we conducted a second iteration incorporating feedback from human reviewers. The main issue identified was that while the rephrased answers might differ from the ground truth, they could still be considered valid. As a result, a new rephrasing iteration was carried out, explicitly prompting the model to account for this nuance, and questions with multiple valid answers were excluded. This led to the removal of 961 samples, leaving the final \careqa{} (open-ended) dataset with 2,769 QA pairs. Consequently, the percentage of correct labels increased to 73.6\%. See Appendix \ref{apx:novel} for further details.

\subsection{Metrics}\label{sec:metrics}

For close-ended evaluations, the metric of choice is accuracy. In contrast, for open-ended queries, there is a variety of metrics which provide different insights into model performance. This work considers eleven of those, which are sorted into four distinct categories:

\begin{itemize}
\setlength\itemsep{0.2em}
    \setlength\parskip{0em}
    \item \textbf{N-gram based metrics} evaluate the overlap of n-grams between the generated and reference answers. This category includes: ROUGE1, ROUGE2, ROUGEL and BLEU.
    \item \textbf{Semantic similarity metrics} evaluate the semantic similarity between the generated text and reference text, often leveraging embeddings or deep learning models. This includes: BERTScore, BLEURT and MoverScore.
    \item \textbf{Perplexity metrics} assess the predictive capabilities of the model by measuring how well it can predict a sequence of words. This includes: Word Perplexity, Bits per Byte and Byte Perplexity. 
    \item \textbf{LLM-judge}: In this category we use the \href{https://huggingface.co/prometheus-eval/prometheus-7b-v2.0}{Prometheus} \cite{kim2024prometheus} model to grade responses based on specific scoring criteria. 

\end{itemize}




\section{Experimentation}






\subsection{Correlation of open-ended vs close-ended}

\begin{figure}[t]
    \centering
    \includegraphics[width=0.46\textwidth]{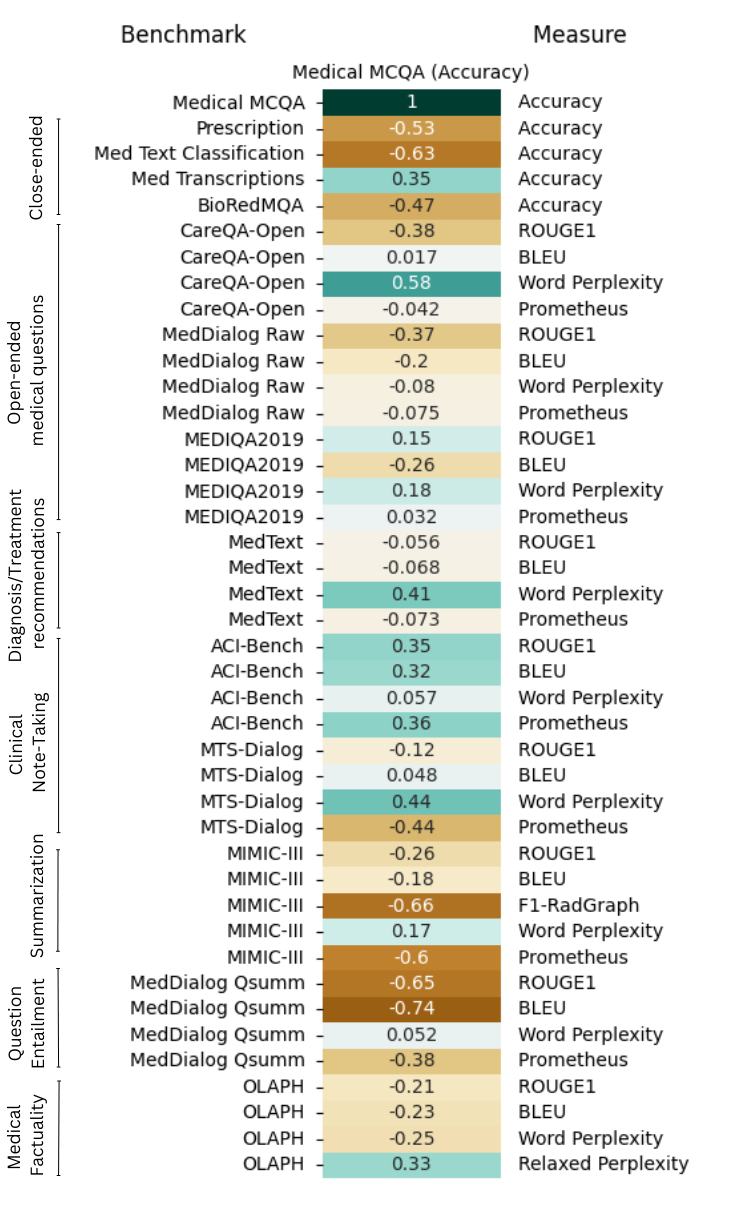} 
    \caption{Correlation between the weighted average accuracy from the MCQA benchmarks and all other close-ended and open-ended tasks and metrics. These results correspond to the smaller models. }
    \label{fig:correlation_type4}
\end{figure}

The first experiment conducted studies the correlation between open-ended and close-ended tasks, as detailed in Table \ref{tab:tasks_bench_metrics}. Specifically, we compare the weighted average accuracy from the various MCQA benchmarks against all other close-ended and open-ended tasks and metrics. Figure \ref{fig:correlation_type4} presents the results for the smaller models.

Of all close and open-ended tasks, only clinical note-taking correlates positively with MCQA, and even in this case, correlation is rather weak. In contrast, summarization, question entailment and the remaining close-ended benchmarks correlate negatively with MCQA, except for Med Transcriptions. The rest show a generalized lack of correlation. The negative correlation could be explained by the lack of medical expertise needed for summarizing and entailing (as information is available in the input), and by the diverse nature of close-ended tasks. At metric level, all open alternatives correlate very weakly with MCQA, except for Perplexity, for which we observe a slight correlation. These findings illustrate the relevance of the benchmarks chosen for evaluation, as well as the complementary nature of MCQA, when considering other tasks like summarization or clinical note-taking. Further details in Appendix \ref{apx:corr_elo}.

\subsection{Correlation of open-ended benchmarks}

The previous section locates open-ended tasks with a variable degree of correlation with close-ended tasks (\eg clinical note-taking, summarization). Let us now analyze correlations within the open-ended category. Details on this are shown in Appendix \ref{apx:metrics_across_bench}.


Notably, no consistently high correlation is observed for any benchmark or task. This suggests that each benchmark measures distinct aspects of model performance. This is the case even for benchmarks tackling the same task (\eg ACI-Bench and MTS-Dialog), illustrating the importance of benchmark source (\ie who crafted the benchmark and in which context). This underscores the need for specialized evaluations for downstream tasks, as generalization cannot be assumed.

\subsection{Correlation of open-ended metrics}

To assess whether the metrics used in the open evaluation are correlated among themselves, and to simplify future analyses for practitioners, we conduct a correlation analysis for each of the metrics detailed in \S\ref{sec:metrics} across all implemented open-ended benchmarks (more details in Appendix \ref{apx:bench_across_metrics}).

This analysis identifies three distinct clusters of highly correlated metrics. The first cluster includes the perplexity metrics, (\ie Word Perplexity, Bits per Byte, and Byte Perplexity)
all of which show a correlation above 0.96 across all analyzed benchmarks. Noticeably, these metrics are all based on probabilistic prediction (perplexity) and information efficiency (
Bits per Byte). The results obtained from Prometheus (an LLM judge) can be considered a distinct cluster of evaluation, illustrating how an external model provides a different and rather unique perspective on model performance. 
Finally, the third cluster includes all n-gram-based metrics, 
together with semantic similarity metrics (\ie BERTScore, BLEURT, and MoverScore). A strong correlation among these metrics is consistently observed across benchmarks, which can be attributed to their shared focus on content and overall text quality.


\subsection{Metrics resilience to rephrasing}

A limitation of open-ended evaluations is their sensitivity to rewording. Let us now analyze the different metrics under this open setup, to better understand their reliability. To do so, the model's output are rephrased, and evaluation recomputed. Six rephrased versions are produced using 
\href{https://huggingface.co/Qwen/Qwen2.5-72B-Instruct}{Qwen2.5-72B-Instruct}. 

Results show that most n-gram-based metrics (\ie ROUGE1, ROUGE2, ROUGEL and BLEU) are resilient to rephrasing. This difference may arise because these metrics rely on surface-level word matching, making them less sensitive to phrasing changes as long as the core vocabulary remains intact. 
\ie in healthcare texts, key terms like `diagnosis,' `treatment,' or medication names often stay consistent, allowing these metrics to maintain a high overlap. In contrast, Prometheus (LLM judge) is the most affected by rewording, which is reasonable considering that, for this evaluation, correct punctuation and formatting in the answers greatly improve scores. This metric is followed by BLEURT and BERTScore (model similarity based) as the least resilient. More details can be found in Appendix \ref{apx:resilience}.

\subsection{Metrics self-consistency}

\begin{figure}[t]
    \centering
    \includegraphics[width=0.46\textwidth]{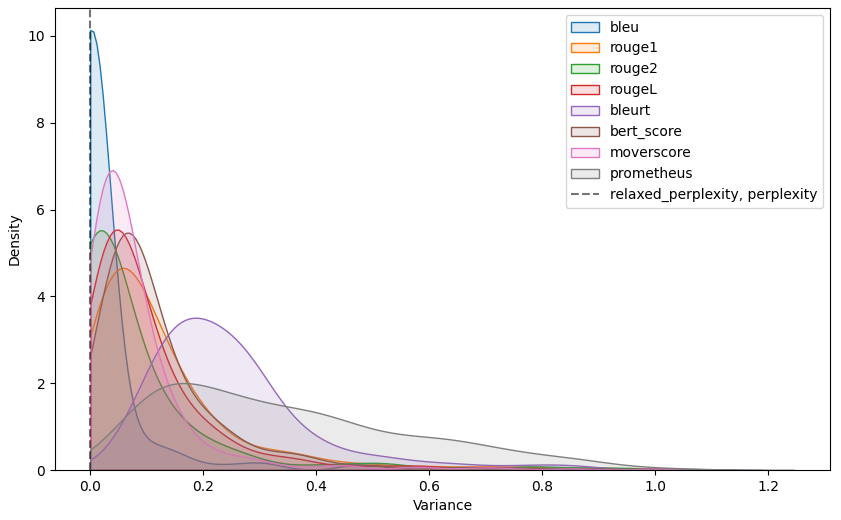}
    \caption{Mean variance distributions across different 
    runs and averaged across models using the \careqa{}-Open dataset. Closer to 0 means more self-consistent.}
    \label{fig:self-consistency}
\end{figure}

Another issue that affects LLM evaluation, particularly on the open-ended setup, is the lack of self-consistency across model runs for some widespread sampling strategies, such as top\_p and top\_k. To evaluate its impact on open-ended evaluation, we generate and evaluate 11 responses for each prompt in \careqa{}-Open using top\_p sampling, $p=0.9$. Results can be seen in Figure \ref{fig:self-consistency}. We observe that among n-gram metrics, BLEU and ROUGE2 are the most self consistent. BLEURT and Prometheus (LLM judge) are the less consistent. Perplexity metrics are perfectly self-consistent. More details can be found in Appendix \ref{apx:self-consistency}.

\section{\relaxed{}: A novel metric}\label{sec:relaxed_perplexity}

By being optimized for next token prediction on the ground truth, LLM's are optimized for perplexity. However, as seen before, this does not necessarily entail good performance on open or close-ended downstream tasks. Additionally, perplexity can be greatly impacted by instruct-tuning and alignment techniques~\cite{lee2024mechanistic}. On the other hand, it has been widely noted that models are more likely to arrive at the correct answer after outputting intermediate tokens, commonly known as chain of thought (CoT) \cite{suzgun2022challenging, wang2023towards}, and that this happens even without specific CoT prompting \cite{wang2024chain}. However, perplexity fails to capture this improvement, and can be negatively impacted by the presence of intermediate tokens.

To evaluate factuality in open-ended benchmarks, with no dependence on confounders or exact formulation while accounting for the potential benefits of intermediate tokens, we propose \relaxed{}. Given a \emph{question} and a \emph{target}, we wish to estimate
\begin{equation*}
    \begin{split}
    \mathbb{P}(target \sim \text{model} \mid question) =  \\
    = \prob(A_0) + \hdots + \prob(A_n \mid B_n)
    \end{split}
\end{equation*}
that is, the probability that the target is sampled from the model given the prompt, at any time in the completion. We denote the events $A_n \equiv \{target \sim \text{model}(question + seq_n)\}$ and $B_n \equiv \{seq_n \sim \text{model}(question)\}$ for any $seq_n$ of $n$ tokens that comes from the model before the target. We can estimate $\prob(A_n \mid B_n)$ as \begin{align*}
    \prob(A_n \mid B_n) \approx \prob(A_n \mid seq^{i_1}_n) + \hdots 
    + \prob(A_n \mid seq^{i_\ell}_n)
\end{align*} for the $\ell$ more likely $n$-token sequences sampled from the model given \emph{question}, because the events $seq^i_n$ and $seq^j_n$ are mutually exclusive. In this notation, $\prob(seq_n^{i_{\ell}}) := \prob(seq_n^{i_{\ell}} \sim \text{model}(question))$. Using this, we can define \relaxed{} as
\begin{equation*}
    \begin{split}
    \text{Relaxed-Perplexity}(target, question, model) = \\
    = \exp\left(-\frac{1}{n + len(target)} \sum_{i=0}^n log P(A_i \mid B_i)\right)
    \end{split}
\end{equation*}
This allows to evaluate correctness in the model's answers probability distribution, with no regard for the exact formulation. Further, for a given prompt and fixed sampling parameters, the metric is perfectly self consistent. We thus test it with the Olaph \cite{jeong2024olaph} medical factuality dataset. In contrast to Perplexity, we observe that \relaxed{} assigns higher scores to models fine-tuned on healthcare datasets. More details on the mathematical formulation, implementation and results of \relaxed{} can be found in Appendix \ref{apx:relaxed_perplexity}.

\section{Conclusions}

This study finds very weak correlations between close-ended and open-ended benchmarks. These results highlight the complementary roles of close-ended and open-ended approaches, and the limited insights provided by individual tests. It thus advocates for broader evaluation setups. Even within open-ended benchmarks targeting the same task (\eg ACI-Bench and MTS-Dialog), no consistently high correlations were found. This indicates that different benchmarks assess distinct model capabilities, underscoring the significance of the benchmark's design.

The analysis of evaluation metrics for open-ended benchmarks identified three distinct clusters that are particularly relevant for assessing medical models: (1) perplexity-based metrics, (2) n-gram-based metrics combined with semantic similarity metrics, and (3) LLM-as-a-judge metrics. Notably, none of these clusters showed strong correlations with the close-ended MCQA evaluation. Additionally, differences in resilience to answer rephrasing and self-consistency were observed, due to the distinct ways these metrics are computed.  


The findings highlight the importance of selecting appropriate benchmarks and evaluation metrics designed for specific tasks. In this regard, the introduced \careqa{} benchmark, featuring both closed- and open-ended formats, serves as a sanity check of existing tests, while the proposed \relaxed{} metric fills a gap in evaluation by focusing on factuality and being resistant to exact formulations in an open-ended setting.



\section{Limitations} 

Since this study is based on specific models, the findings may not generalize to other LLM architectures. Additionally, the quality and diversity of the datasets used for evaluation are limited, meaning these benchmarks may not fully capture the performance of LLMs across the broader healthcare landscape. While metrics and benchmarks can indicate how well LLMs perform on certain tasks, they may not reflect the complexities of integrating LLMs into real-world healthcare practices. 

In evaluating the models, we observed that applying the model’s chat template to MCQA tasks led to decreased performance, whereas open-ended evaluations showed improvement. To ensure a fair comparison between open-ended and MCQA evaluations, we maintained the same configuration across both categories and did not apply the model's chat template to any of the evaluations.

Regarding the new benchmark introduced, although subject matter experts created the original exam materials, which underwent public scrutiny, \careqa{} has not been subjected to formal bias assessment. Consequently, it may not adequately represent the full spectrum of medical knowledge or encompass all possible patient demographics. Furthermore, although a human review was performed on the open-ended version, it has not undergone thorough evaluation by healthcare experts, raising the possibility of errors or biases introduced by the LLM used to rephrase the questions. Therefore, we advise users to exercise caution when interpreting and generalizing the results.

All experiments are conducted on English benchmarks  (except for the Spanish version of \careqa{}), and generalization to other languages has not been considered. To enable reproducibility, all resources are made available. \careqa{} is accessible on Hugging Face\footnote{\url{https://huggingface.co/datasets/HPAI-BSC/CareQA}} and all new tasks are accessible in the original \textit{lm-evaluation-harness} framework\footnote{\url{https://github.com/EleutherAI/lm-evaluation-harness}}.

\section*{Acknowledgements}
This work is supported by Anna Arias Duart, Pablo Agustin Martin Torres and Daniel Hinjos García fellowships within the “Generación D” initiative, \href{https://www.red.es/es}{Red.es}, Ministerio para la Transformación Digital y de la Función Pública, for talent atraction (C005/24-ED CV1). Funded by the European Union NextGenerationEU funds, through PRTR.

We also acknowledge the computational resources provided by the FinisTerrae III, Leonardo, and MareNostrum 5 supercomputers. We are particularly grateful to the Operations department at BSC for their technical support.

Lastly, we sincerely thank Jordi Bayarri-Planas, Atia Cortés, Orlando Montenegro and Òscar Molina for their valuable time and feedback during the human evaluation process.

\bibliography{custom}

\appendix
\include{latex/appendix_novel_bench}

\input{latex/appendix_correlations}

\input{latex/appendix_resilience}

\input{latex/appendix_relaxed_perplexity}



\input{latex/appendix_results}

\label{sec:perplexity}

\end{document}

%% file: latex/tables/table_5.tex
\begin{table*}[t]
\centering
\resizebox{0.92\textwidth}{!}{%
\begin{tabular}{rcl}
\multicolumn{3}{c}{{\color[HTML]{000000} \Large \textsc{Close-ended}}}                                                                                     \\ \hline 
{\color[HTML]{000000} \textbf{\large \textsc{Tasks}}}                                                                       & {\color[HTML]{000000} \textbf{\large \textsc{Metrics}}}                                                                                  & {\color[HTML]{000000} \textbf{\large \textsc{Datasets}}}                                                                                                                                                                                                      \\ \hline
\rowcolor[HTML]{EFEFD2} 
{\color[HTML]{656565} \textbf{\begin{tabular}[c]{@{}r@{}}Multiple choice\\ questions\end{tabular}}}   & {\color[HTML]{656565} Accuracy}                                                                                          & {\color[HTML]{656565} \begin{tabular}[c]{@{}l@{}}· MedMCQA \cite{pmlr-v174-pal22a} \hspace{30pt} · PubMedQA \cite{jin2019pubmedqa} \\ · MedQA \cite{jin2020disease} \hspace{41pt} · MMLU \cite{hendrycks2020measuring}    \\ · \href{https://huggingface.co/datasets/HPAI-BSC/CareQA}{\careqa{}-Close} \end{tabular}}                                                                                   \\
\rowcolor[HTML]{E3EFD6} 
{\color[HTML]{656565} \textbf{\begin{tabular}[c]{@{}r@{}}Prescriptions \\ writing \end{tabular}}}            & {\color[HTML]{656565} "}                                                                                          & {\color[HTML]{656565} · \href{https://huggingface.co/datasets/devlocalhost/prescription-full}{Prescription} }                                                                                                                                                                                \\

\rowcolor[HTML]{C8E9D9} 
{\color[HTML]{656565} \textbf{\begin{tabular}[c]{@{}r@{}}Medical text\\ classification\end{tabular}}}      & {\color[HTML]{656565} "}                                                                                          & {\color[HTML]{656565} \begin{tabular}[c]{@{}l@{}}· \href{https://www.kaggle.com/datasets/chaitanyakck/medical-text/data}{Medical Text for classification} \cite{10.1145/3582768.3582795}\\ · \href{https://www.kaggle.com/datasets/tboyle10/medicaltranscriptions}{Medical Transcriptions}\end{tabular}} \\

\rowcolor[HTML]{CDEBEB} 
{\color[HTML]{656565} \textbf{\begin{tabular}[c]{@{}r@{}}Relation\\ extraction\end{tabular}}}              & {\color[HTML]{656565} "}                                                                                          & {\color[HTML]{656565} · \href{https://huggingface.co/datasets/YufeiHFUT/BioRED_all_info}{BioRED} \cite{luo2022biored}}         \\

\\ \hline
\multicolumn{3}{l}{}                                                                                                                                                                                                                                           \\ 
\multicolumn{3}{c}{{\color[HTML]{000000} \Large \textsc{Open-ended}}}                                                                                                                                                                                          \\ \hline

\rowcolor[HTML]{C7DBE9} 
        
{\color[HTML]{656565} \textbf{\begin{tabular}[c]{@{}r@{}}Open-ended\\ medical questions\end{tabular}}}     & {\color[HTML]{656565} \normalsize \begin{tabular}[c]{@{}c@{}}BLEU, BLEURT, ROUGE,\\ BERTScore, MoverScore,\\ Prometheus, Perplexity\end{tabular}} & {\color[HTML]{656565} \begin{tabular}[c]{@{}l@{}}· \href{https://huggingface.co/datasets/bigbio/meddialog}{MedDialog Raw} \cite{zeng2020meddialog} \\ · \href{https://huggingface.co/datasets/bigbio/mediqa_qa}{MEDIQA2019} \cite{MEDIQA2019} \\· \href{https://huggingface.co/datasets/HPAI-BSC/CareQA}{\careqa{}-Open} 
\end{tabular}}

\\

\rowcolor[HTML]{E1DEE9} 
{\color[HTML]{656565} \textbf{\begin{tabular}[c]{@{}r@{}}Making diagnosis \\ and treatment\\ recommendations\end{tabular}}} & {\color[HTML]{656565} " }  & {\color[HTML]{656565} · \href{https://huggingface.co/datasets/BI55/MedText}{MedText} }                                                                                                                                                                                      
\\

\rowcolor[HTML]{E5CFDF} 
{\color[HTML]{656565} \textbf{\begin{tabular}[c]{@{}r@{}}Clinical\\ note-taking\end{tabular}}}             & {\color[HTML]{656565} " } & {\color[HTML]{656565} \begin{tabular}[c]{@{}l@{}}· \href{https://huggingface.co/datasets/har1/MTS_Dialogue-Clinical_Note}{MTS-Dialog} \cite{mts-dialog}\\ · ACI-Bench \cite{aci-bench}\end{tabular} }                                                                            

\\ 

\rowcolor[HTML]{EFD6DC} 
{\color[HTML]{656565} \textbf{\begin{tabular}[c]{@{}r@{}}Medical\\ factuality\end{tabular}}}             & {\color[HTML]{656565} \begin{tabular}[c]{@{}c@{}} " \\ + Relaxed Perplexity \end{tabular}  } & {\color[HTML]{656565} \begin{tabular}[c]{@{}l@{}} · \href{https://huggingface.co/datasets/dmis-lab/MedLFQA}{OLAPH} \cite{jeong2024olaph}\end{tabular} }                                                                                                     \\ 

\rowcolor[HTML]{E9E1DE} 
{\color[HTML]{656565} \textbf{Summarization}}                                                              & {\color[HTML]{656565} \begin{tabular}[c]{@{}c@{}} " \\ + F1-RadGraph \end{tabular}} & {\color[HTML]{656565} · \href{https://huggingface.co/datasets/dmacres/mimiciii-hospitalcourse-meta}{MIMIC-III} \cite{johnson2016mimic}}                                                                                                                                              \\

\rowcolor[HTML]{E4D9D4} 
{\color[HTML]{656565} \textbf{\begin{tabular}[c]{@{}r@{}}Question\\ entailment\end{tabular}}}              & {\color[HTML]{656565} "}  & {\color[HTML]{656565} · \href{https://huggingface.co/datasets/lighteval/med_dialog}{Meddialog Qsumm} \cite{zeng2020meddialog} }                                                               

\\ \hline

\end{tabular}
}\caption{This table presents the tasks implemented in this paper. The first column specifies the different tasks. The second details the metrics used (ROUGE includes ROUGE1, ROUGE2 and ROUGEL, and Perplexity includes Bits per Byte, Byte Perplexity, and Word Perplexity). The third column outlines the benchmarks used for each task.} \label{tab:tasks_bench_metrics}
\end{table*}

%% file: latex/appendix_novel_bench.tex
\section{Novel Benchmarks}\label{apx:novel}

\subsection{\careqa{} (close-ended)}\label{apx:careqa-close}

\careqa{} is a novel benchmark for evaluating healthcare Large Language Models (LLMs) through multiple-choice question answering. \careqa{} was created by collecting exam materials in PDF format from the official Spanish government website. These documents were automatically parsed and then underwent post-processing to ensure data quality. This process involved removing 23 inaccurately parsed instances and excluding officially impugned questions. To enhance global accessibility, the original Spanish questions were translated into English using GPT-4. 

\begin{figure}[h]
    \centering
    \includegraphics[width=0.49\textwidth]{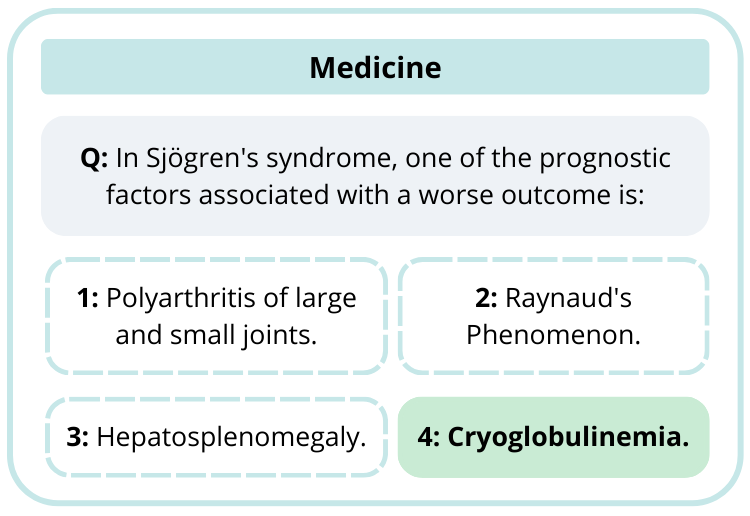}
    \caption{\careqa{} example from Medicine category.}
    \label{fig:careqa_example_1}
    \vspace{10px}
\end{figure}

Each \careqa{} sample contains metadata including a numeric exam identifier, full question text, four answer options, correct answer, exam year, and specialization category. The dataset is available in both Spanish and English, facilitating cross-lingual research. Examples of \careqa{} samples are provided in Figure~\ref{fig:careqa_example_1} and Table~\ref{tab:careqa_examples_1}.

\begin{figure*}[t]
    \centering
    \includegraphics[width=1.0\textwidth]{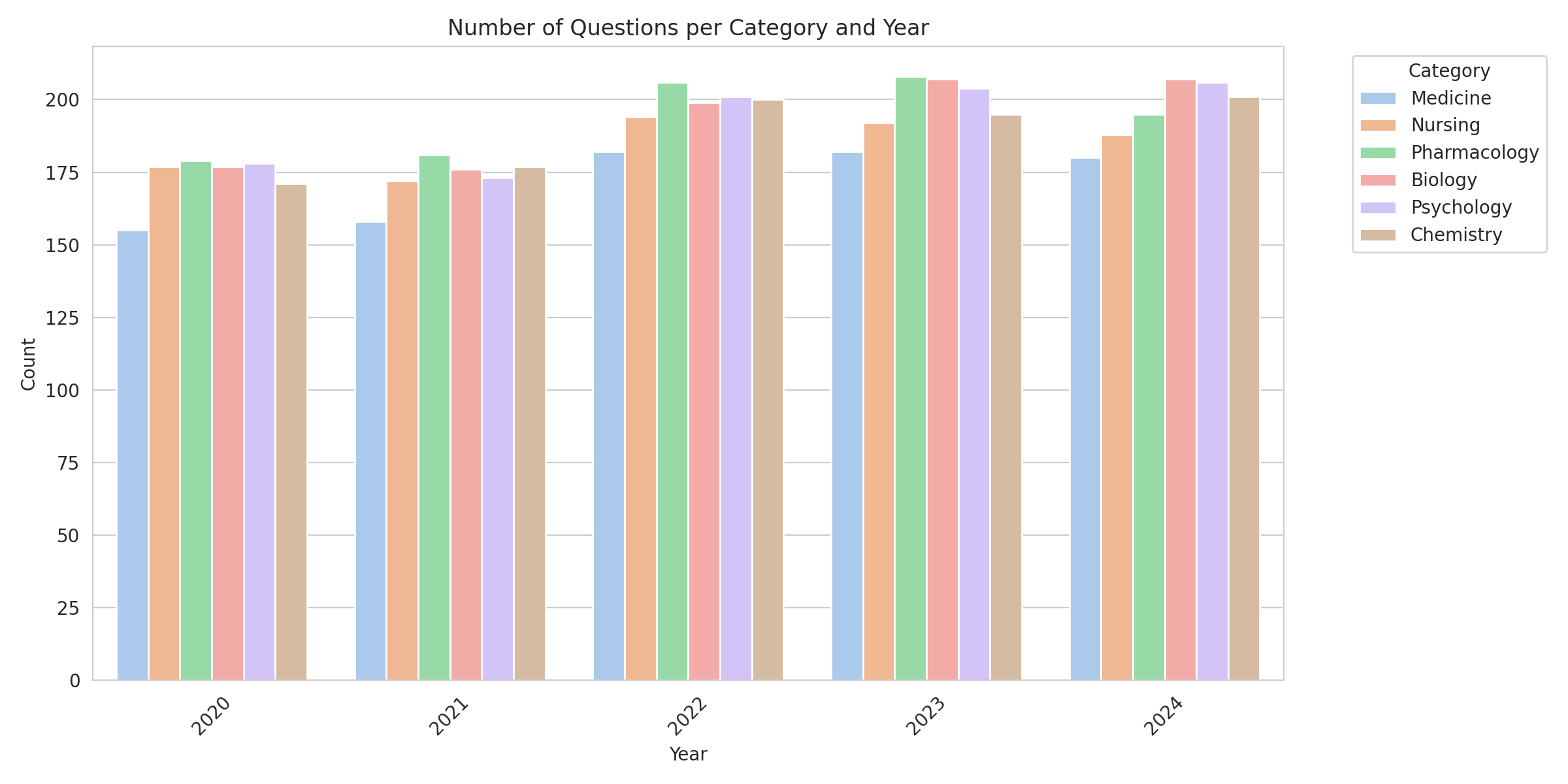}
    \caption{Category distribution per Category and Year (\careqa{} close-ended)}
    \label{fig:careqa_categories_yearly}
\end{figure*}

While \careqa{} shares its source with HeadQA in the Spanish Specialised Healthcare Training (MIR) exams, there is no overlap between the datasets. \careqa{} expands upon its predecessor, covering the years 2020 to 2024 and comprising 5,621 question-answer test pairs, compared to HeadQA's 2,742 test pairs from 2013 to 2017.
The dataset's composition is illustrated in Figure~\ref{fig:careqa_categories_yearly}, showing the category distribution by year to reveal potential temporal trends in exam content.


Table~\ref{tab:careqa_stats} presents additional information about the dataset, including the total number of questions per category, the longest and average question and answer lengths (in tokens), and the overall vocabulary size. This comprehensive overview of \careqa{}'s structure and content demonstrates its potential as a valuable resource for evaluating and improving healthcare-focused language models.

\subsection{\careqa{} (open-ended)}\label{apx:careqa-open}
We developed the open-ended dataset by adapting the existing closed-ended \careqa{} dataset through the expansion of the English set. The first step was to filter out questions that contained terms such as "incorrect", "except", "false", "not correct", or "NOT", as these terms indicate that the questions focus on identifying incorrect answers among the provided options. After this filtering, we rephrased the remaining questions into an open-ended format using the \href{https://huggingface.co/Qwen/Qwen2.5-72B-Instruct}{Qwen2.5-72B-Instruct} model, specifically instructing it to only rephrase questions that could be effectively transformed. This process excluded questions that explicitly ask for incorrect options or require a selection from the provided answers. We employed two different prompts for rephrasing, followed by a selection process to determine the best-rephrased version or to discard the question if neither was suitable. 

Initially, the close-ended \careqa{} contained 5,621 QA pairs, but after the rephrasing process, the number of suitable questions for the open-ended version was reduced to 3,730 QA pairs. This new dataset retains the same categories as the closed-ended version, including medicine, nursing, biology, chemistry, psychology, and pharmacology.

\begin{figure*}[t]
    \centering
    \subfloat[]{
        \includegraphics[width=1.0\textwidth]{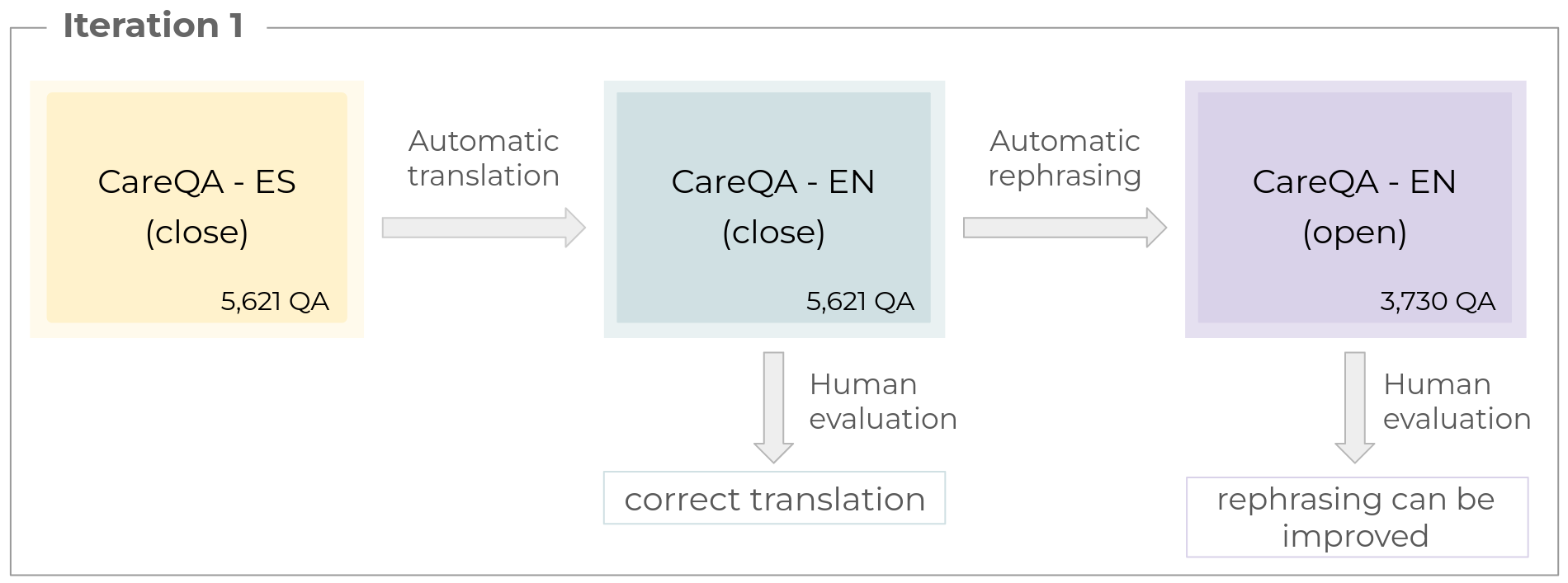}
        \label{fig:iterations-1}
    }
    \vspace{1em} 
    \subfloat[]{
        \includegraphics[width=1.0\textwidth]{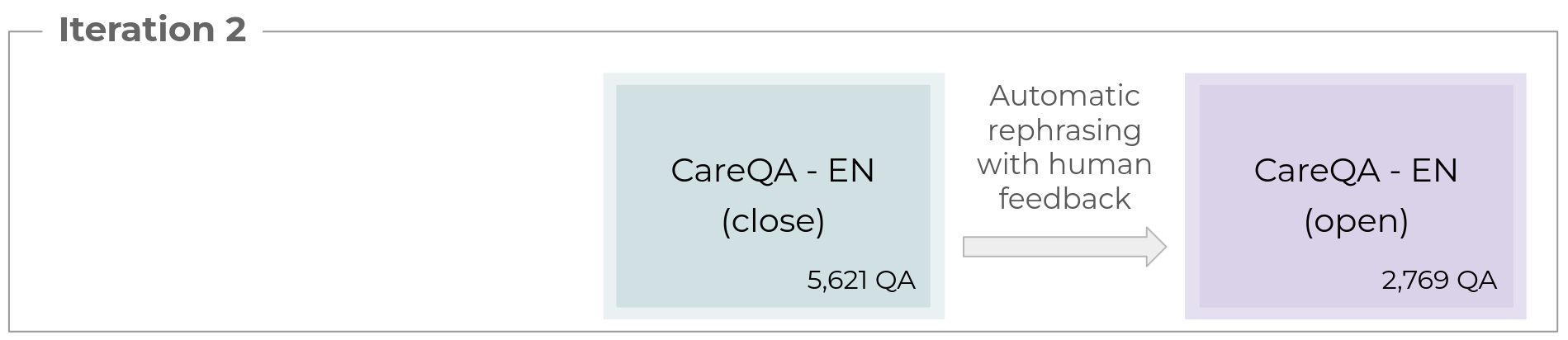}
        \label{fig:iterations-2}
    }
    \caption{Iterations with human evaluators to create the \careqa{} dataset in English, including both open and closed versions.}
    \label{fig:iterations}
\end{figure*}

Based on feedback from the human review (detailed in \S\ref{sec:human_eval}), a second iteration of rephrasing was conducted, as illustrated in Figure \ref{fig:iterations}. In this phase, the model was instructed to validate only questions that could be answered exclusively using the ground truth, ensuring there were no alternative correct answers. As a result, 961 questions were removed, reducing the \careqa{} (open-ended) dataset to a total of 2,769 QA pairs. 

\begin{figure*}[t]
    \centering
    \includegraphics[width=1.0\textwidth]{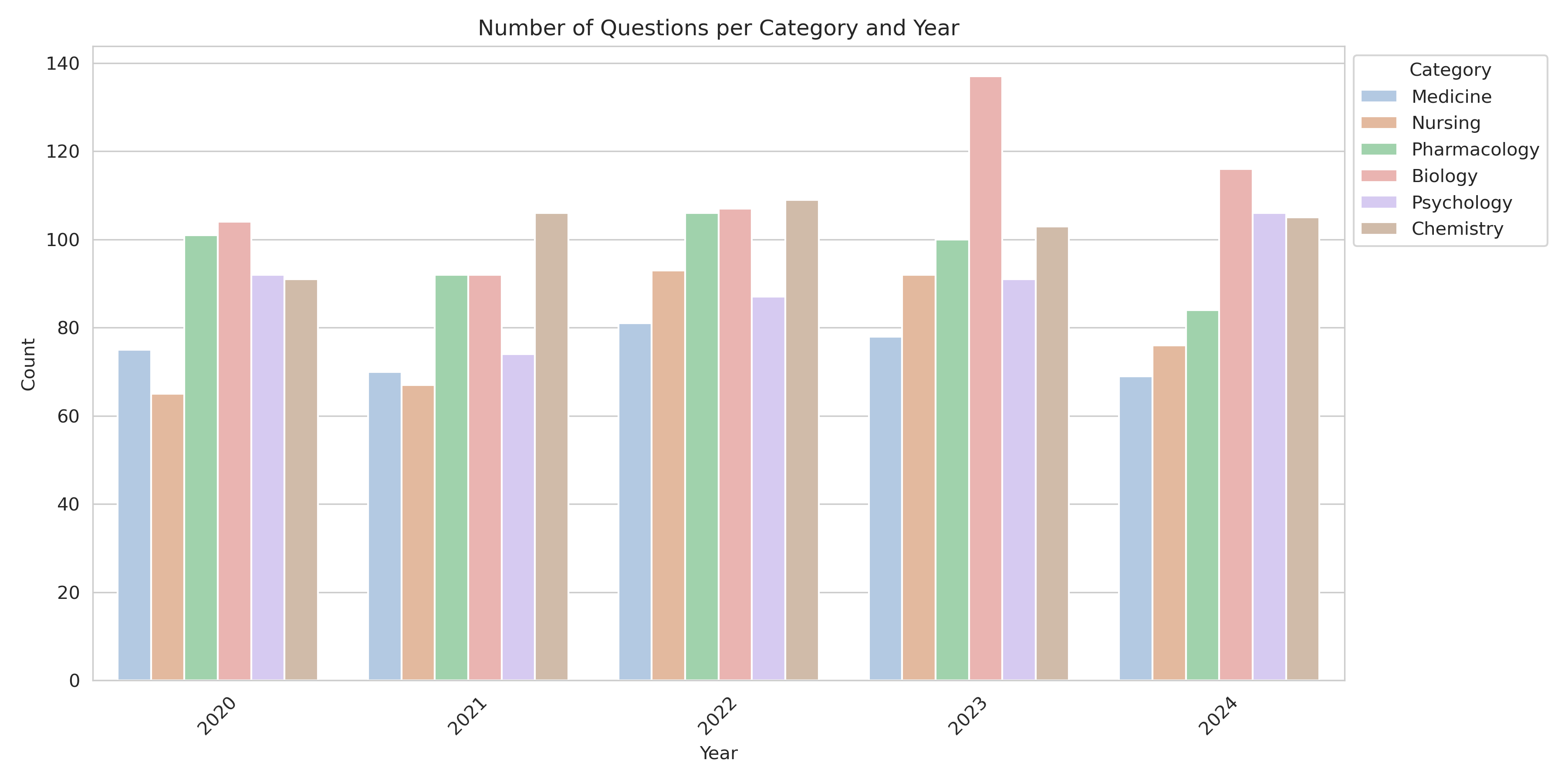}
    \caption{Category distribution per Category and Year (\careqa{} open-ended).}
    \label{fig:careqa_categories_yearly_open}
\end{figure*}

Figure \ref{fig:careqa_categories_yearly_open} illustrates the distribution of these 2,769 QA pairs in the open-ended version and examples of QA pairs from both the close-ended and open-ended versions of the \careqa{} dataset are shown in Table \ref{tab:careqa_examples}. Both datasets are publicly available\footnote{\url{https://huggingface.co/datasets/HPAI-BSC/CareQA}}.

\subsection{Human evaluation}\label{sec:human_eval}

To validate the translations performed by GPT-4 for the English version of \careqa, as well as the rephrasing process executed by Qwen2.5-72B-Instruct for the open-ended \careqa, a human evaluation was conducted with 10 human evaluators, including 5 authors of this article.

We selected a total of 260 QA pairs for evaluation, covering both translation and rephrasing. This sample size ensures a confidence level of 95\% with a margin of error of 5\% for translation and 5.73\% for rephrasing. Each question was evaluated by at least three evaluators. 

\input{latex/tables/careqa_human_results_iter1}

The results are shown in Table \ref{tab:evaluation_results_iter1} and correspond to the percentages of correct answers labeled by at least one evaluator, by two evaluators, and by all three evaluators. For both translation and rephrasing, the percentage of questions labeled as correct by at least one evaluator is high (98.6\% for translation and 96.1\% for rephrasing). However, when considering the cases where all three evaluators agreed on the correctness of the QA pair, the percentages drop: 83.1\% for translation and 65.8\% for rephrasing (first iteration).

\begin{figure*}[t]
    \centering
    \includegraphics[width=1.0\textwidth]{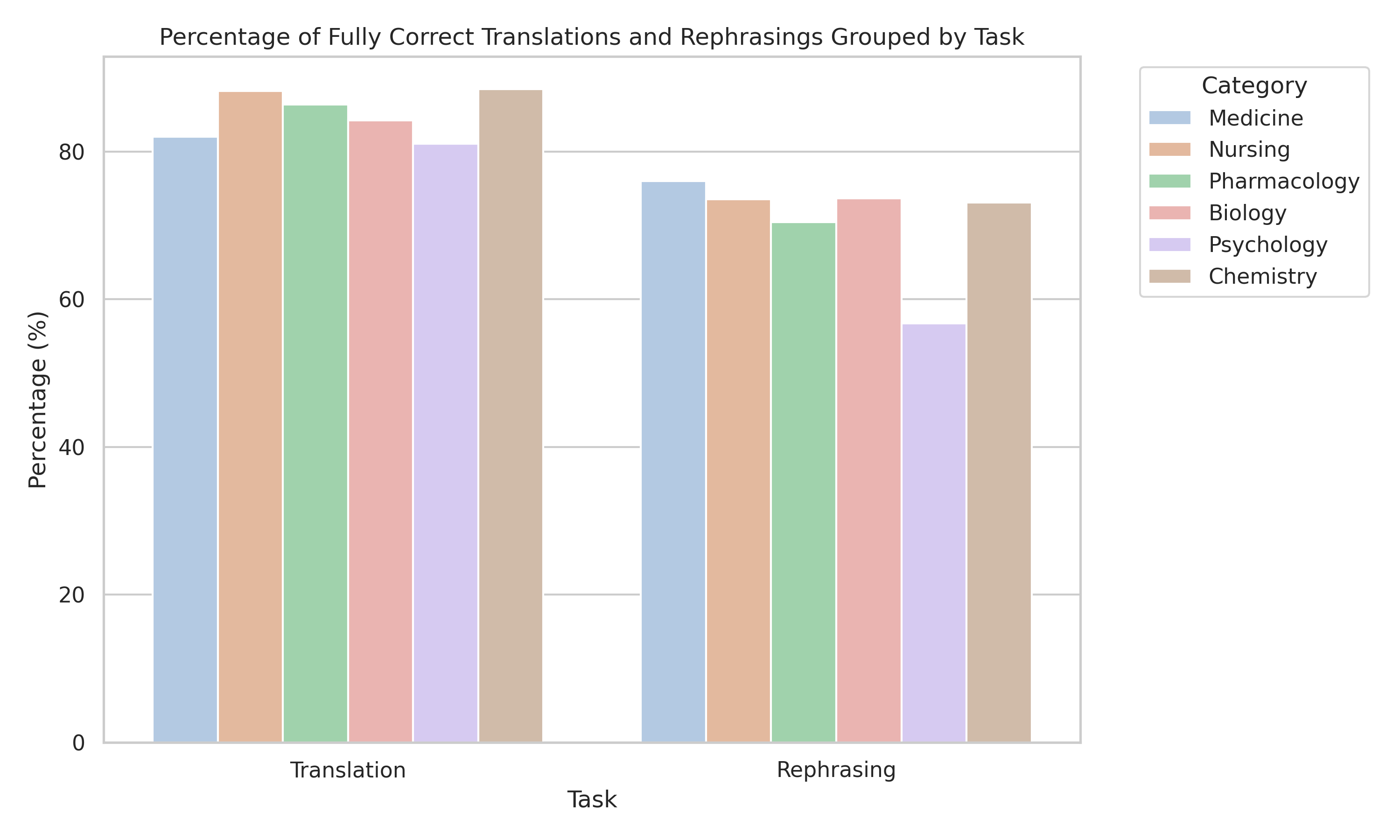}
    \caption{Correctness distribution per Category \careqa{} (open-ended).}
    \label{fig:careqa_categories_correct}
\end{figure*}

For translation, the agreement percentage was considered sufficiently high, and the English dataset was deemed valid. In contrast, for the open-ended rephrasing version, the agreement rate was not high enough, so a second iteration of rephrasing, as explained in the previous section, was carried out. After removing invalid questions, the percentage of correct answers increased, see third column of Table \ref{tab:evaluation_results_iter1}. After this second iteration, the open dataset was also considered valid. The final agreement of both tasks grouped per category can be seen in Figure \ref{fig:careqa_categories_correct}.

\input{latex/careqa_examples}

\input{latex/tables/careqa_stats}

\input{latex/tables/careqa_open_table}

%% file: latex/tables/careqa_human_results_iter1.tex
\begin{table}[h]
\centering
\resizebox{0.99\linewidth}{!}{%
\begin{tabular}{lccc}
\hline
\textbf{Agreement} & \textbf{Translation (\%)} & \multicolumn{2}{c}{\textbf{Rephrasing (\%)}} \\ 
                    &                      & \textbf{Iter 1 }  & \textbf{Iter 2}  \\ 
\hline
Correct (1/3)      & 98.6                      & 96.1      &  98.1   \\ 
Correct (2/3)      & 96.7                      & 85.8      &   92.8   \\ 
Correct (3/3)      & 83.1                      & 65.8    &   73.6       \\ 
Interrater & 84.4                      & 69.7        &    75.5    \\ 
\hline
\end{tabular}%
}
\caption{Evaluation results for translation and rephrasing. The first row shows the percentage of correct samples tagged by at least one evaluator. The second row refers to samples tagged as correct by two evaluators. The third row indicates samples labeled as correct by all three evaluators. The last row shows the agreement rate among the three evaluators.}
\label{tab:evaluation_results_iter1}
\end{table}

%% file: latex/careqa_examples.tex
\begin{table*}[t]
\centering
\renewcommand{\arraystretch}{1.5}
\resizebox{2\columnwidth}{!}{%
\begin{tabular}{lllllll}
\hline
\textbf{Question} & \textbf{Option 1} & \textbf{Option 2} & \textbf{Option 3} & \textbf{Option 3} & \textbf{Year} & \textbf{Category} \\ \hline
The Glisson's capsule covers: & Spleen. & \textbf{Liver.} & Kidney. & Lung. & 2024 & Biology \\
Cardiolipin is a: & Sphingolipid. & \textbf{Phosphoglyceride.} & Steroid. & Ganglioside. & 2020 & Biology \\
The cinnamic acid is a: & Terpene. & Fatty acid. & Flavonoid. & \textbf{Phenylpropanoid.} & 2021 & Chemistry \\
Which of the following acids is strongest?: & HCl. & \textbf{HI.} & H2SO4. & HNO3. & 2023 & Chemistry \\
Indicate the ketogenic amino acid: & Cysteine. & Glutamine. & Methionine. & \textbf{Lysine.} & 2020 & Pharmacology \\
O2 and O3 are examples of: & Isotopes. & \textbf{Allotropes.} & Isomers. & Conformers. & 2023 & Pharmacology \\
Malignant hyperthermia is not related to: & Succinylcholine. & Desflurane. & \textbf{Propofol.} & Sevoflurane. & 2024 & Medicine \\
The most common benign tumors of the esophagus are: & Fibrovascular polyps. & \textbf{The leiomyomas.} & Squamous papillomas. & The hemangiomas. & 2021 & Medicine \\
Which opioid presents a higher analgesic potency? & Morphine. & Methadone. & Meperidine. & \textbf{Fentanyl.} & 2023 & Nursing \\
Indicate the antidote for ethylene glycol: & Methylene blue. & \textbf{Fomepizole.} & Carnitine. & Dimercaprol. & 2024 & Nursing \\
Olfactory hallucinations are more common in: & Delirium. & Manic episode. & \textbf{Epilepsy.} & Alcoholic hallucinosis. & 2022 & Psychology \\
What kind of drug is quetiapine? & A benzodiazepine. & An anxiolytic. & An antidepressant. & \textbf{An antipsychotic.} & 2020 & Psychology \\
\hline
\end{tabular}
}
\caption{Examples of \careqa{} (close-ended) samples. Correct options are marked in bold. Questions were selected based on length for space reasons.} 
\label{tab:careqa_examples_1}
\end{table*}

%% file: latex/tables/careqa_stats.tex
\begin{table*}[ht]
\centering
\renewcommand{\arraystretch}{1.5}
\resizebox{2\columnwidth}{!}{%
\begin{tabular}{ccccccc}
& 
&  
& 
\textbf{\careqa{}
} &
& \\
\hline
&  
\textbf{QA Pairs} &
\textbf{Max Q tokens} & 
\textbf{Avg Q tokens} & 
\textbf{Max A tokens} & 
\textbf{Avg A tokens} &
\textbf{Vocab} \\ 
\hline
Medicine   
& 857
& 202
& 48.57
& 43 
& 9.65 
& 9626 \\
Nursing
& 923
& 96
& 24.61
& 70
& 12
& 9113 \\
Pharmacology
& 969
& 147
& 18.94 
& 56
& 8.51
& 7906 \\
Biology
& 966
& 51
& 12.82
& 48
& 6.6 
& 6300 \\
Psychology
& 962
& 208
& 22.60
& 67
& 9.92 
& 7573 \\
Chemistry
& 944
& 81
& 16.88
& 47 
& 8.2 
& 6022 \\
\hline
\end{tabular}
} 
\vspace{5px}
\caption{\careqa{} (close-ended) dataset statistics, where Q and A represents the Question and Answer respectively.}
\label{tab:careqa_stats}
\vspace{5px}
\end{table*}

%% file: latex/tables/careqa_open_table.tex
\begin{table*}[h]
\renewcommand{\arraystretch}{1.5}
    \resizebox{2\columnwidth}{!}{%

\begin{tabular}{@{}rlll@{}}
\toprule
\multicolumn{1}{l}{} & \multicolumn{1}{c}{\textbf{Close-ended}} & \multicolumn{1}{c}{\textbf{Open-ended}} & \textbf{Category} \\ \midrule
\textbf{Question} & \begin{tabular}[c]{@{}l@{}}The best way to estimate the relative strength of hydrogen\\ bonds between the molecules of halogen hydrides, H-X, is \\ by measuring:\end{tabular} & \begin{tabular}[c]{@{}l@{}}What is the best way to estimate the relative strength of\\ hydrogen bonds between the molecules of halogen\\ hydrides, H-X?\end{tabular} & \multirow{2}{*}{Chemistry} \\ \cmidrule(lr){2-3}
\textbf{Answer} & The enthalpies of vaporization & The enthalpies of vaporization. &  \\ \midrule
\textbf{Question} & \begin{tabular}[c]{@{}l@{}}Taking into account the general principles regarding the\\ minimum interval between the non-simultaneous\\ administration of vaccines, identify the minimum interval\\  between 2 attenuated vaccines:\end{tabular} & \begin{tabular}[c]{@{}l@{}}What is the minimum interval recommended between\\  the non-simultaneous administration of two attenuated \\ vaccines, according to general principles?\end{tabular} & \multirow{2}{*}{Nursing} \\ \cmidrule(lr){2-3}
\textbf{Answer} & Four weeks. & Four weeks. &  \\ \midrule
\textbf{Question} & \begin{tabular}[c]{@{}l@{}}We evaluated in the emergency room an adult person who\\ is irritable, yawning, complaining of muscle pain and\\  cramps. They are nauseous and have notable tearing. \\ The pupils are dilated. Which of the following is the \\ most probable diagnosis?\end{tabular} & \begin{tabular}[c]{@{}l@{}}An adult patient presents to the emergency room with\\ irritability, yawning, muscle pain and cramps, nausea, \\ notable tearing, and dilated pupils. What is the most \\ probable diagnosis based on these symptoms?\end{tabular} & \multirow{2}{*}{Medicine} \\ \cmidrule(r){1-3}
\textbf{Answer} & Opioid abstinence. & Opioid abstinence. &  \\ \bottomrule
\end{tabular}

}\caption{Examples of QA pairs: On the left, the close-ended version from \careqa{}, and on the right, the open-ended version.} \label{tab:careqa_examples}
\end{table*}

%% file: latex/appendix_correlations.tex
\section{Correlations}\label{apx:correlations}

\subsection{Correlations between MCQA and Elo results}\label{apx:corr_elo}

We perform a correlation analysis on the performance results of the medical MCQA benchmarks listed in Table \ref{tab:tasks_bench_metrics}. Additionally, we include Elo scores from the Chatbot Arena\footnote{\url{https://lmarena.ai/}}, a crowdsourcing platform that collects pairs of model-generated answers in response to user prompts, where the user selects the winning model based on their criteria.

\begin{figure}[t]
        \centering
        \includegraphics[width=0.48\textwidth]{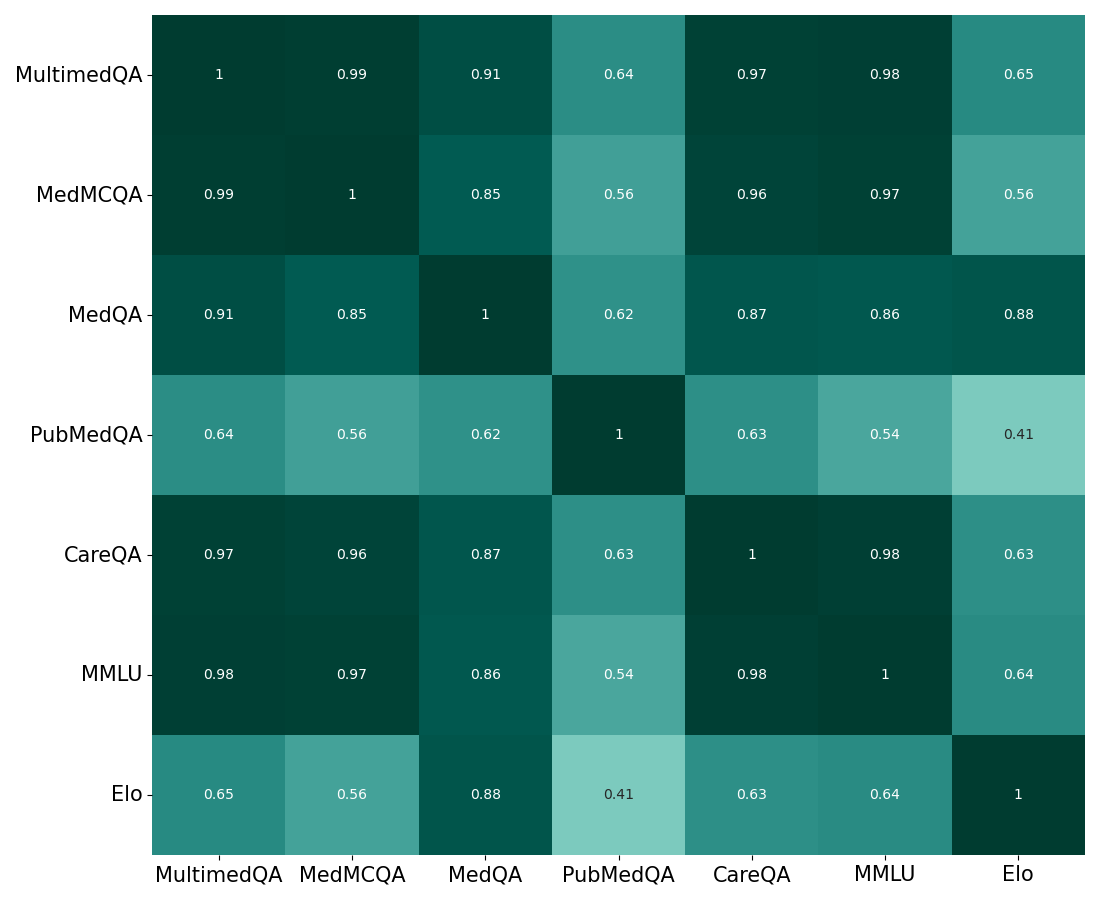}
        \caption{Comparison of correlations between MCQA benchmarks and ELO results for small models.}
        \label{fig:corr_small}
    \end{figure}

\begin{figure}[h] 
        \centering

        \includegraphics[width=0.48\textwidth]{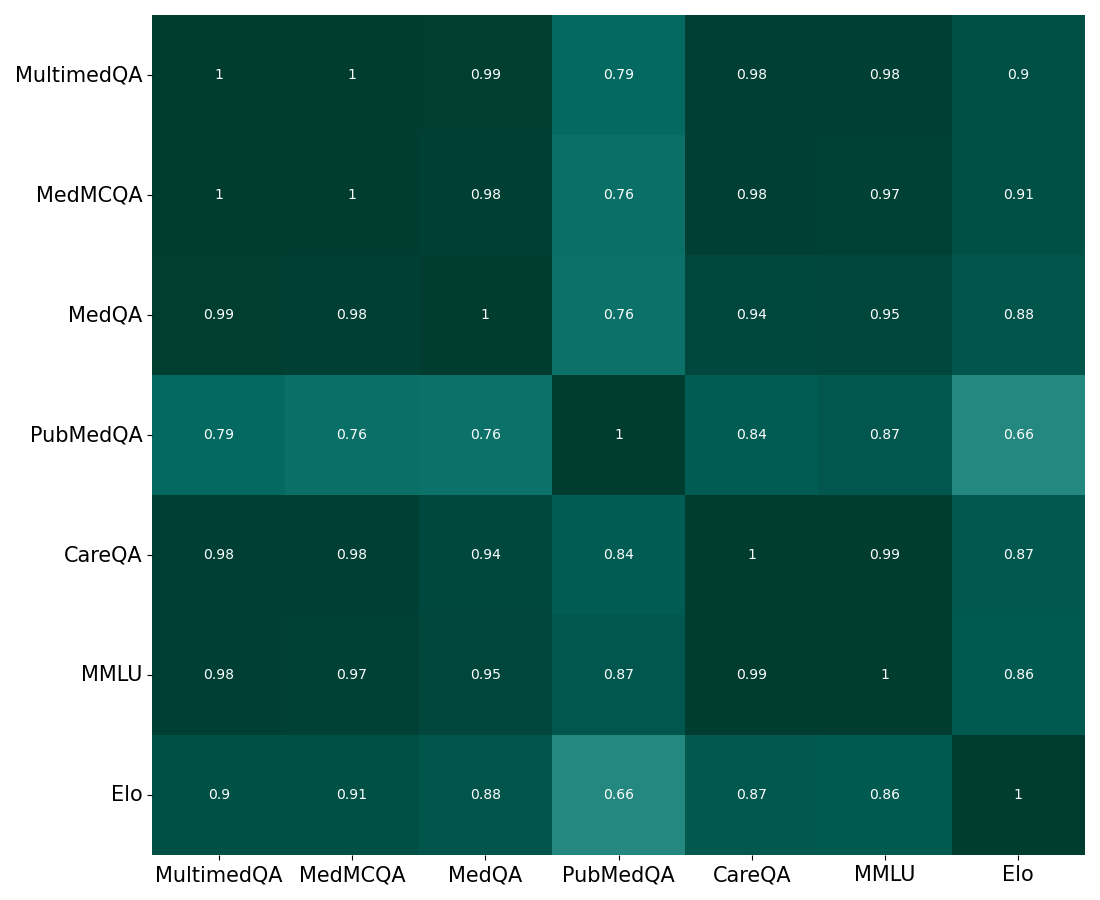}
        \caption{Comparison of correlations between MCQA benchmarks and ELO results for medium models.}
        \label{fig:corr_medium}
\end{figure}

We conducted a correlation analysis using both small and medium models. The small models used for the correlation shown in Figure \ref{fig:corr_small} are as follows: \href{https://huggingface.co/google/gemma-2-9b-it}{gemma-2-9b-it} \cite{gemma_2024}, \href{https://huggingface.co/meta-llama/Meta-Llama-3-8B-Instruct}{Meta-Llama-3.1-8B-Instruct}\cite{llama3modelcard},
\href{https://huggingface.co/mistralai/Mistral-7B-Instruct-v0.2}{Mistral-7B-Instruct-v0.2}, 
\href{https://huggingface.co/mistralai/Mistral-7B-Instruct-v0.3}{Mistral-7B-Instruct-v0.3}, 
\href{https://huggingface.co/microsoft/Phi-3-mini-4k-instruct}{Phi-3-mini-4k-instruct}, 
\href{https://huggingface.co/microsoft/Phi-3-medium-4k-instruct}{Phi-3-medium-4k-instruct},
\href{https://huggingface.co/Qwen/Qwen1.5-7B-Chat}{Qwen1.5-7B-Chat},
\href{https://huggingface.co/unsloth/Starling-LM-7B-beta}{Starling-LM-7B-beta},
\href{https://huggingface.co/unsloth/Starling-LM-7B-beta}{Starling-LM-7B-beta} and
\href{https://huggingface.co/01-ai/Yi-1.5-9B-Chat}{Yi-1.5-9B-Chat}. And the medium models used in Figure \ref{fig:corr_medium} are as follows: \href{https://huggingface.co/Nexusflow/Athene-70B}{Athene-70B}\cite{Athene2024},
\href{https://huggingface.co/allenai/tulu-2-dpo-70b}{tulu-2-dpo-70b}\cite{ivison2023camels},
\href{https://huggingface.co/01-ai/Yi-1.5-34B-Chat}{Yi-1.5-34B-Chat},
\href{https://huggingface.co/google/gemma-2-27b-it}{gemma-2-27b-it},
\href{https://huggingface.co/meta-llama/Llama-3.1-70B-Instruct}{Llama-3.1-70B-Instruct},
\href{https://huggingface.co/mistralai/Mixtral-8x7B-Instruct-v0.1}{Mixtral-8x7B-Instruct-v0.1},
\href{https://huggingface.co/Qwen/Qwen2-72B-Instruct}{Qwen2-72B-Instruct}\cite{qwen2},
and \href{https://huggingface.co/WizardLMTeam/WizardLM-70B-V1.0}{WizardLM-70B-V1.0}

From this analysis, we found that MedQA, MedMCQA, \careqa{}, and MMLU are highly correlated with one another. However, PubMedQA exhibits a noticeably lower correlation with the other medical benchmarks, particularly in smaller models.

Regarding the Elo scores, we observe a moderate correlation with the MCQA benchmarks, with the correlation being significantly stronger for larger models. This is likely due to larger models' ability to produce more coherent responses. Non-expert evaluators, such as those in the Elo scoring system, may favor responses that are well-structured and fluent, even if they lack precise medical accuracy. As a result, this preference for more polished answers could lead to a higher correlation with MCQA performance.



    \hfill
    
\clearpage
\subsection{Correlation between metrics}\label{apx:bench_across_metrics}

\begin{figure}[t]
    \centering
    \includegraphics[width=0.48\textwidth]{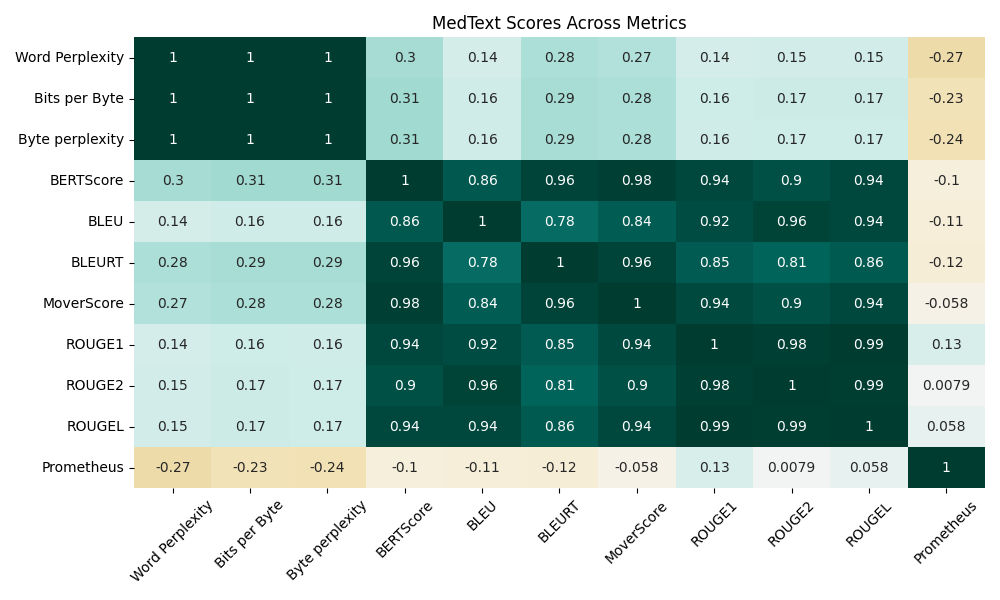}
    \caption{This correlation matrix illustrates the relationships among the different open-ended metrics used to evaluate the benchmark for diagnosis and treatment recommendations. Three distinct clusters of metrics are identified: (1) perplexity metrics, (2) n-gram and semantic similarity metrics, and (3) Prometheus metrics.}
    \label{fig:medtext_corr}
\end{figure}

In this correlation analysis, we fix the open-ended benchmark and examine the correlations across the various computed metrics. Figure \ref{fig:medtext_corr}, presents the correlation matrix for the benchmark focused on making diagnosis and treatment recommendations, highlighting the three clusters of metrics identified in the paper. This correlation matrix was also computed for the rest of benchmarks revealing three similar clusters. The matrices were computed using the following models: \href{https://huggingface.co/BioMistral/BioMistral-MedMNX}{BioMistral-MedMNX},  \href{https://huggingface.co/johnsnowlabs/JSL-MedLlama-3-8B-v2.0}{JSL-MedLlama-3-8B-v2.0}, \href{https://huggingface.co/microsoft/Phi-3-mini-4k-instruct}{Phi-3-mini-4k-instruct}, \href{https://huggingface.co/mistralai/Mistral-7B-Instruct-v0.3}{Mistral-7B-Instruct-v0.3}, \href{https://huggingface.co/Qwen/Qwen2-7B-Instruct}{Qwen2-7B-Instruct} \cite{qwen2}, \href{https://huggingface.co/m42-health/Llama3-Med42-8B}{Llama3-Med42-8B} \cite{christophe2024med42}, \href{https://huggingface.co/meta-llama/Meta-Llama-3-8B-Instruct}{Meta-Llama-3.1-8B-Instruct}\cite{llama3modelcard} \href{https://huggingface.co/01-ai/Yi-1.5-9B-Chat}{Yi-1.5-9B-Chat} \cite{young2024yi}, \href{https://huggingface.co/microsoft/Phi-3-medium-4k-instruct}{Phi-3-medium-4k-instruct}, \href{https://huggingface.co/01-ai/Yi-1.5-34B-Chat}{Yi-1.5-34B-Chat} \cite{young2024yi}, \href{https://huggingface.co/mistralai/Mixtral-8x7B-Instruct-v0.1}{Mixtral-8x7B-Instruct-v0.1}.

\subsection{Correlations of benchmarks} \label{apx:metrics_across_bench}

\begin{figure}[t]
    \centering
    \includegraphics[width=0.48\textwidth]{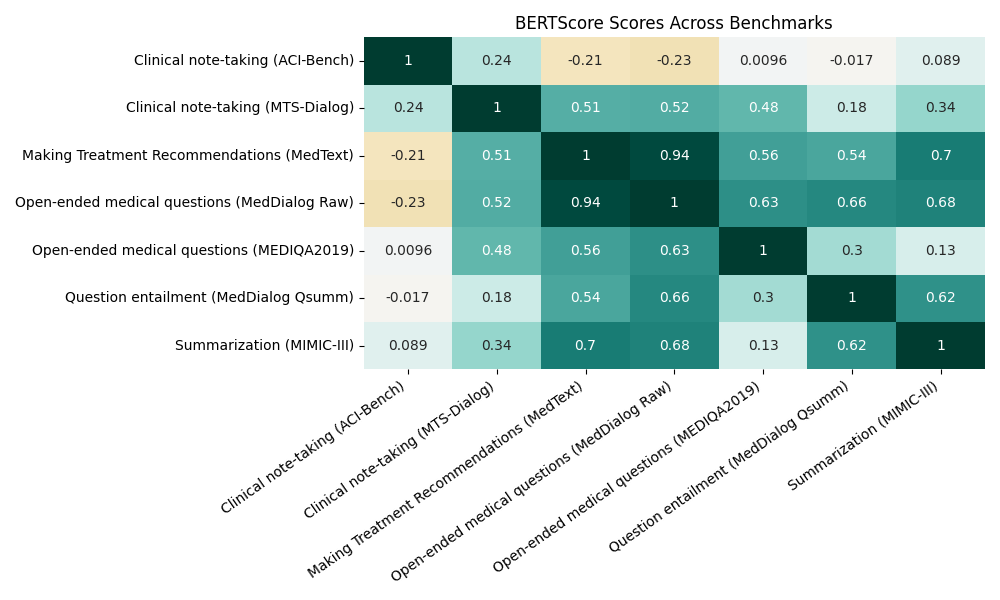}
    \caption{Correlations of BERTScore across benchmarks.}
    \label{fig:corr_bert_acros_bench}
\end{figure}

\begin{figure}[t]
    \centering
    \includegraphics[width=0.48\textwidth]{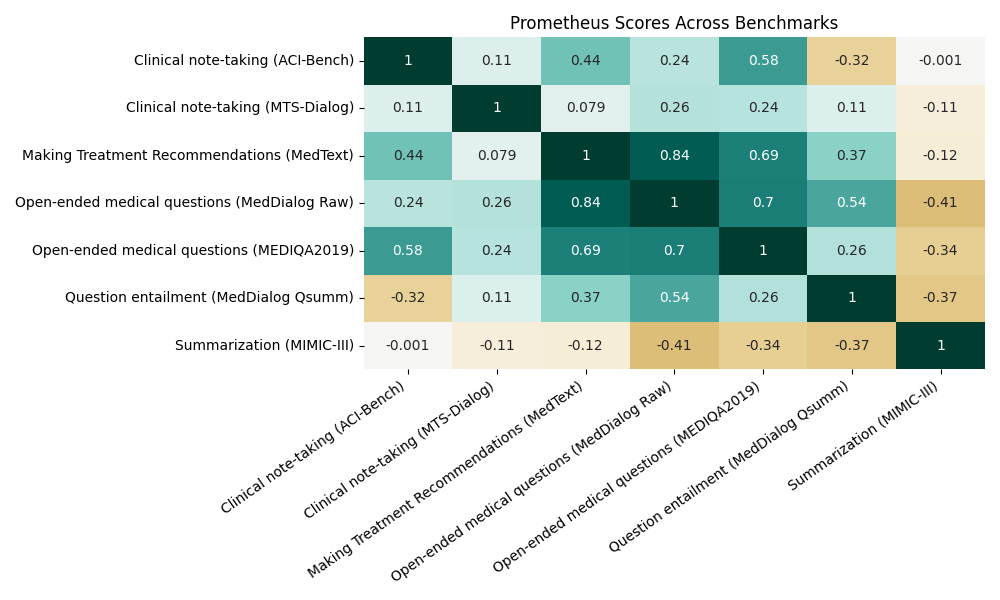}
    \caption{Correlation of Prometheus scores across benchmarks.}
    \label{fig:corr_prometheus}
    \vspace{10px}
\end{figure}

In this correlation analysis we study the relationships between specific metrics across all the open-ended benchmarks implemented. As stated in the paper, no consistent high correlation was observed among all metrics for any benchmark or task. Examples of these correlation matrices are shown in Figures \ref{fig:corr_bert_acros_bench} and \ref{fig:corr_prometheus}. The models used to generate these correlation matrices are the same as those described in the Appendix \ref{apx:bench_across_metrics}.

%% file: latex/appendix_resilience.tex
\section{Resilience to rephrasing and self-consistency}\label{apx:resilience_and_consistency}

\begin{figure}[t]
    \centering
    \includegraphics[width=0.48\textwidth]{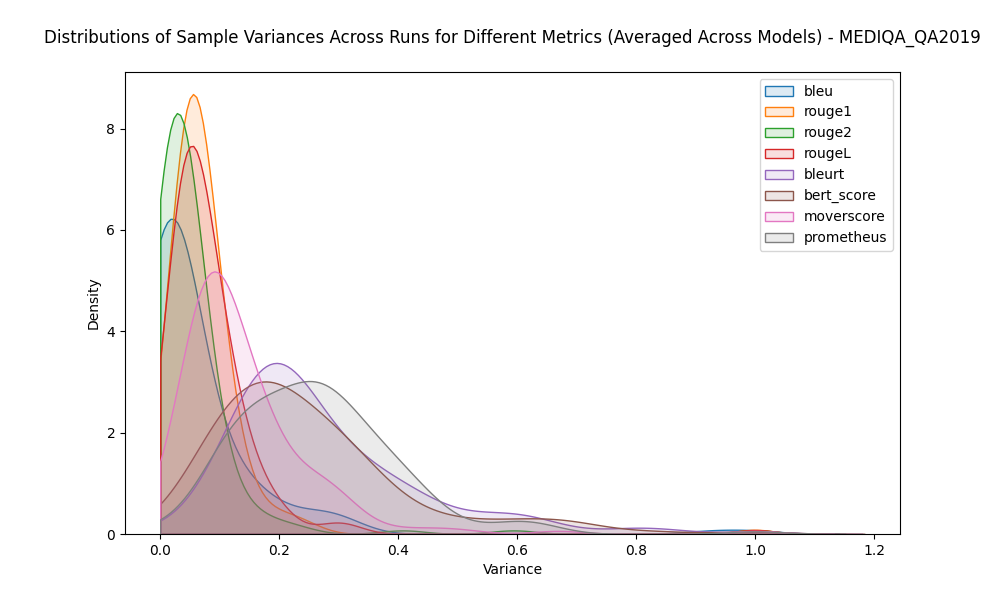}
    \caption{Mean variance distributions across different rephrasings and models using the MEDIQA2019 dataset. Each metric is represented by a different color.}
    \label{fig:resilience_1}
\end{figure}

\begin{figure}[t]
    \centering
    \includegraphics[width=0.48\textwidth]{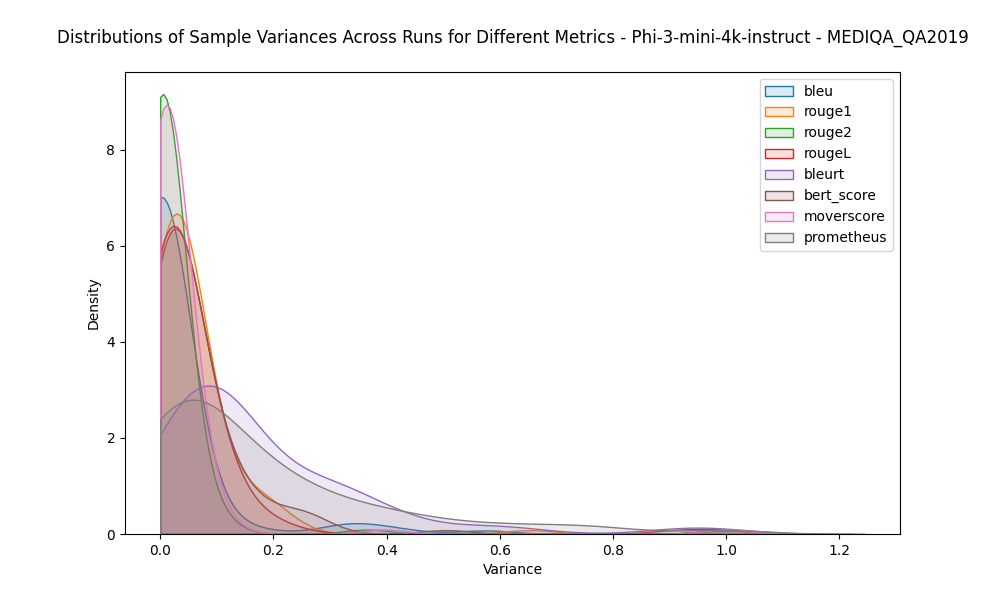}
    \caption{Mean variance distributions across different rephrasings using the Phi-3-mini-4k-instruct model and the MEDIQA2019 dataset. Each metric is represented by a different color.}
    \label{fig:resilience_2}
\end{figure}

\begin{figure}[t]
    \centering
    \includegraphics[width=0.49\textwidth]{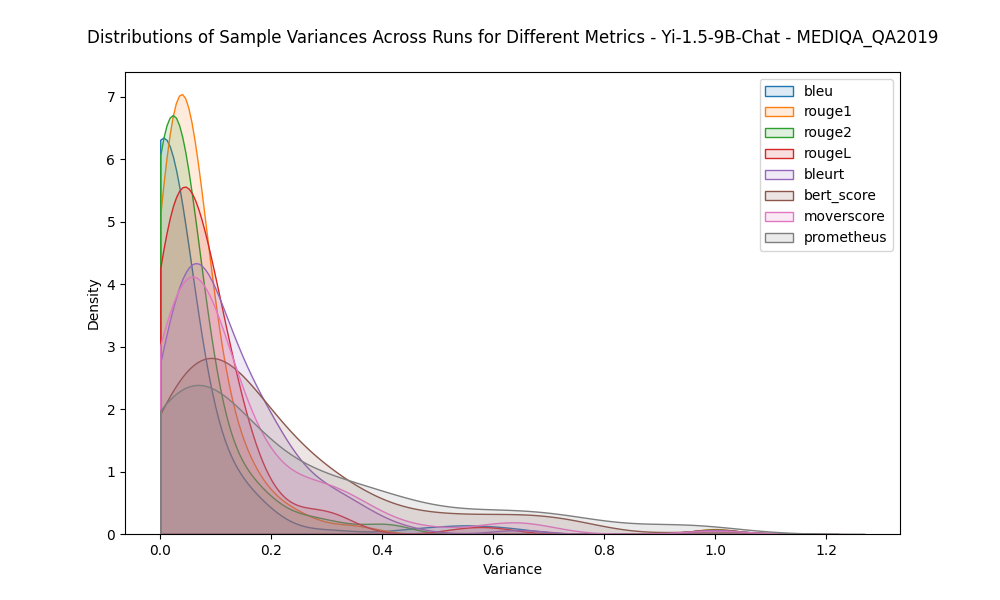}
    \caption{Mean variance distributions across different rephrasings using the Yi-1.5-9B-Chat model and the MEDIQA2019 dataset. Each metric is represented by a different color.}
    \label{fig:resilience_3}
\end{figure}

\subsection{Resilience}\label{apx:resilience}

As described earlier, we conducted this experiment by rephrasing the model outputs six times and re-computing the metrics. We used both Qwen2.5-72B-Instruct and Meta-Llama-70B-Instruct with the following \textit{system\_prompt}: ``You are a helpful rephrasing assistant. Rephrase the prompt provided without changing its original meaning, but do not try to address or answer it in any case."

We run the script 5 times on recorded model answers with top\_p sampling to obtain several rephrasings of each answer. After manual inspection, the outputs of Qwen2.5-72B-Instruct were deemed of higher quality.

Figure \ref{fig:resilience_1} shows the mean variance across all runs for the MEDIQA2019 dataset. Before plotting, we scale variances by dividing by the max interval (max value - min value) in each column. Figures \ref{fig:resilience_2} and \ref{fig:resilience_3} present the variance distributions for two specific models. Figure \ref{fig:resilience_2} displays the results for the Phi-3-mini-4k-instruct model, while Figure \ref{fig:resilience_3} shows the results for the Yi-1.5-9B-Chat model.

In Figure \ref{fig:resilience_1} we can observe three different clusters: rouge metrics (low mean-variance, low meta-variance), bleu and moverscore (low mean-variance, medium meta-variance) and bert\_score, bleurt, prometheus (high mean variance, high meta-variance).

\subsection{Self-consistency}\label{apx:self-consistency}

\begin{figure}[h]
    \centering
    \includegraphics[width=0.48\textwidth]{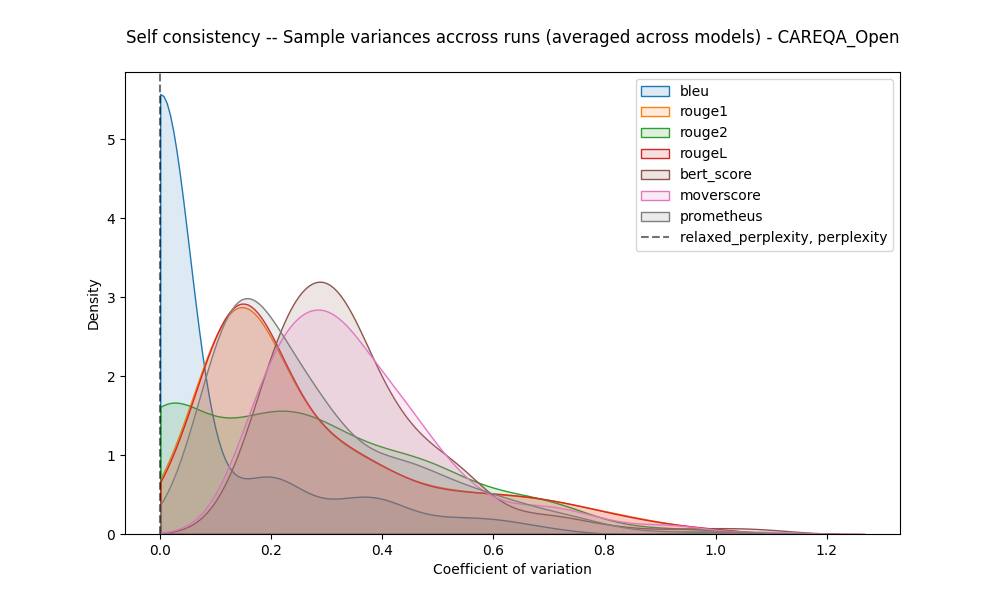}
    \caption{Mean coefficient of variation distributions across different runs and averaged across models for self-consistency. Each metric is represented by a different color.}
    \label{fig:self-consistency_cv}
\end{figure}

As described earlier, we conducted this experiment by prompting models with each question in \careqa{}-Open for a number of repetitions ($r$). We fix $r=11$. Sampling parameters used where $\text{top\_p} = 0.9$ and $\text{temperature} = 1$. We compute variances per prompt, and then average across models. Results can be seen in Figure \ref{fig:self-consistency}. Besides, we compute the coefficient of variation, defined for prompt $p$ as:
\begin{align*}
    CV(p) = \frac{1}{\mu_p}\sqrt{\frac{\sum_i(x_i - \mu_p)^2}{N}}
\end{align*}
Then we average across models, and plot the $CV$ distribution for all prompts in \careqa{}-Open. Results can be seen in Figure \ref{fig:self-consistency_cv}. From this computation we remove the BLEURT metric, for it can take negative values.

%% file: latex/appendix_relaxed_perplexity.tex
\section{Novel Metric: \relaxed{}} \label{apx:relaxed_perplexity}

As mentioned before, we define \relaxed{} as
\begin{equation*}
\begin{split}
    \text{Relaxed-Perplexity}(target, question, model) = \\
    = \exp\left(-\frac{1}{n + len(target)} \sum_{i=0}^n log P(A_i \mid B_i)\right)
\end{split}
\end{equation*}
for events \[A_n \equiv \{target \sim \text{model}(question + seq_n)\}\] and \[B_n \equiv \{seq_n \sim \text{model}(question)\}.\] That is, $A_n$ is the event that target is sampled from the model inputted with $question + seq_n$, for any $seq_n$ of $n$ tokens.

\begin{figure}[t]
    \centering
    \includegraphics[width=0.43\textwidth]{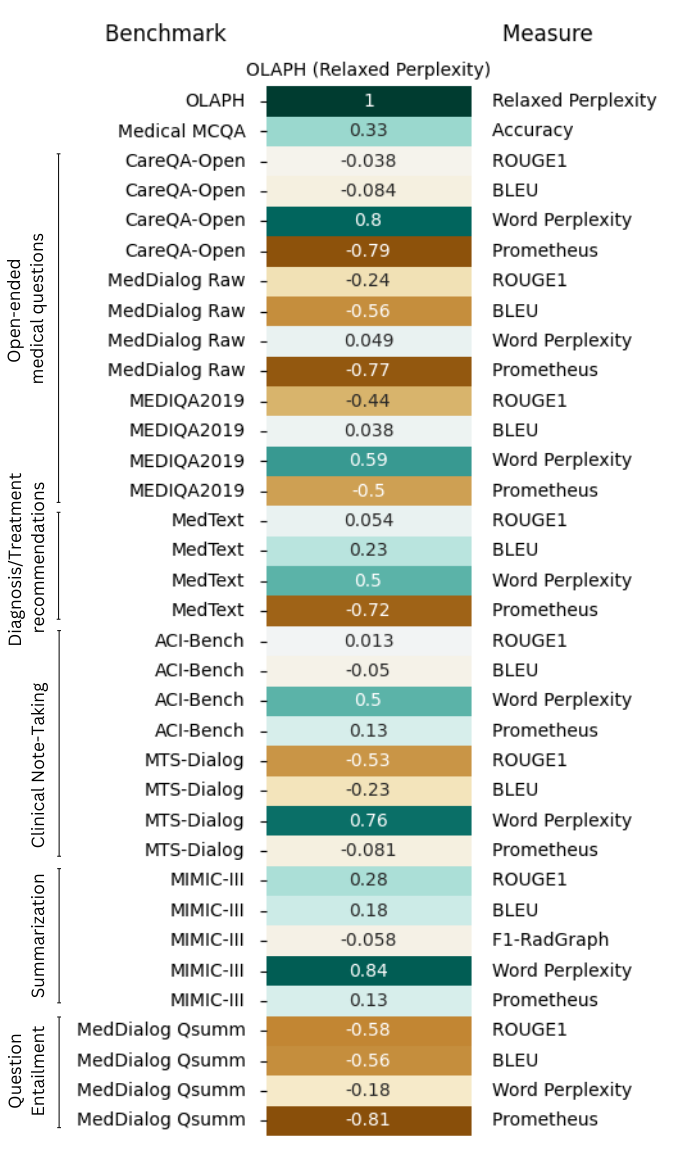}
    \caption{Correlation between OLAPH - \relaxed{} and the rest of benchmarks.}
    \label{fig:correlations_olaph}
\end{figure}

Thus, in order to compute $\prob(A_n \mid B_n)$ we need to take into account the probability distribution of all $n$-token model answers when the input is $question$, which is extremely costly (with computational time exponential in $n$). 
In fact, by the law of total probability we would have
\begin{align*}
    \prob(A_n \mid B_n)~\prob(B_n) = \prob(A_n \mid seq^1_n)~\prob(seq^1_n) + \cdots \\
+ \prob(A_n \mid seq^{q^n}_n)~\prob(seq^{q^n}_n)
\end{align*}

$q$ being the size of the vocabulary. This holds because the events $seq^i_n$ and $seq^j_n$ are mutually exclusive. In this notation, $\prob(seq_n^{i_{\ell}}) := \prob(seq_n^{i_{\ell}} \sim \text{model}(question))$, and also $\prob(B_n) = ~\prob(\cup_{i} seq_n^{i})$.

However, given that almost all this combinations of tokens contribute with negligible probabilities to the sum, we can estimate the above quantity as
\begin{align*}
    \prob(A_n \mid B_n) \approx \prob(A_n \mid seq^{i_1}_n)~\prob(seq^{i_1}_n) + \hdots \\
+ \prob(A_n \mid seq^{i_\ell}_n)~\prob(seq^{i_\ell}_n)
\end{align*}
for the $\ell$ more likely $n$-token sequences sampled from the model given \emph{question}, which can be computed efficiently using beam search, diverse beam search \cite{vijayakumar2016diverse} or top\_p sampling. 

Notice that also $\prob(B_n) = 1$ unless stop tokens appeared before in the completion, and then the value decreases for big $n$. In our implementation, where $max\_tokens \in [128, 256]$, stop tokens rarely appear and so we estimate $\prob(B_n) \approx 1$.

Now, there is an issue with this formulation. We noticed that, since $\prob(seq_n^{i})$ is the joint probability of all tokens in the sequence, as $n$ grows this value collapses very quickly. In fact, among the $\ell$ most likely sequences, we may bound 
\begin{align*}
    \frac{1}{c_n} \leq \prob(seq_n^i) \leq \frac{1}{d_n}
\end{align*}
for constants $c_n$ and $d_n$ that only depend on $n$ (for example, take the average and max prob of sequences of that length respectively; also, notice $d_n \leq n$ ). And thus we may take
\begin{align*}
    \prob(A_n \mid B_n) &\approx \\ \frac{c_n + d_n}{2 c_n d_n} \left( \prob(A_n \mid seq^{i_1}_n) + \hdots + \prob(A_n \mid seq^{i_\ell}_n) \right)
\end{align*}

This effectively assigns more value to the target appearing earlier in the completion, benefiting models that do not verbose and biasing comparisons without adding real value, for this constant does not depend on the target. In order to deal with this, we \emph{skew} the models distribution with respect to length by multiplying with the inverse of the constant, and end up with the final approximation:
\begin{align*}
    \prob(A_n \mid B_n) \approx \prob(A_n \mid seq^{i_1}_n) + \hdots + \prob(A_n \mid seq^{i_\ell}_n)
\end{align*}
Notice this step may be omitted depending on the evaluation goal.

\relaxed{} is specifically designed to evaluate factuality in the answers, with no regard for the exact formulation. We thus test it with the OLAPH \cite{jeong2024olaph} dataset, and note that for more effective evaluation of other open-ended benchmarks, some preprocessing of the ground truths must be carried out.

For our experiments we use top-p sampling, selecting the $\ell \in \{5, 10\}$ best sentences in a search space of $s \in \{10, 100\}$. We observe similar results with all combinations, and so fix $\ell = 5$ and $s = 10$ for better performance.

\begin{figure*}[t]
    \centering
    \includegraphics[width=0.88\textwidth]{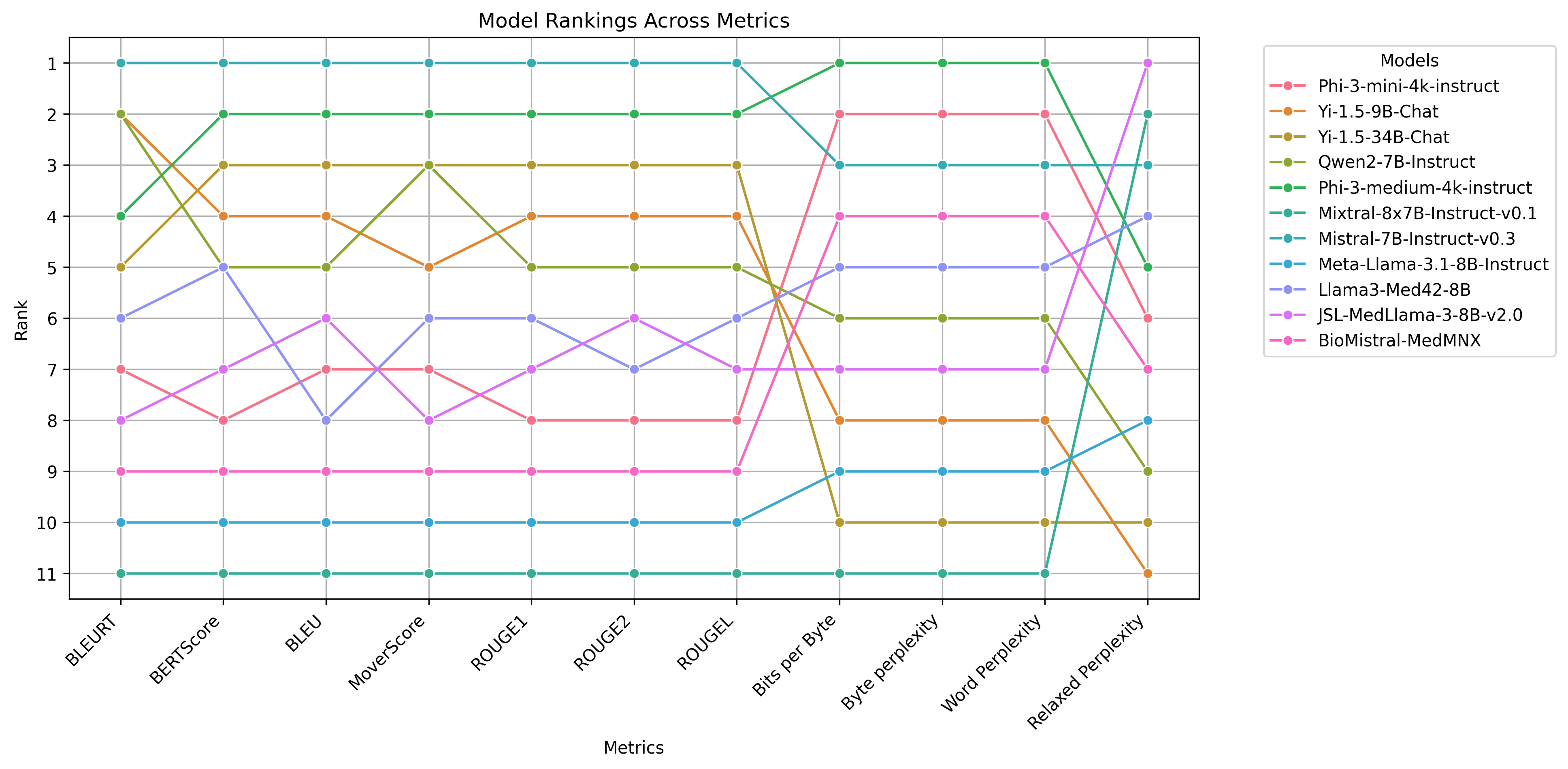}
    \caption{  
    Ranking results for all models on the OLAPH medical factuality dataset for all metrics. The top position is ranked as 1 and the lowest as 11. Different models are represented in distinct colors. It can be seen there is low agreement across metrics.
    }
    \label{fig:ranking}
\end{figure*}

\input{latex/tables/olaph_examples}

We add another hyperparameter, which we denote as \emph{stride}, for better efficiency. Instead of computing \( \sum_{i=0}^n log P(A_i \mid B_i)\) we compute \( \sum_{i=0,i+stride}^n log P(A_i \mid B_i)\), which we find to be as effective. We select $stride \in \{8, 16\}$.

The implementation is built using vllm\footnote{\href{https://github.com/vllm-project/vllm}{https://github.com/vllm-project/vllm}}, which provides tools for efficient LLM inference \cite{kwon2023efficient}. It remains as future work to implement \relaxed{} with beam search.

\subsection{Connection with cross-entropy}

The exponent of perplexities can be understood as a cross-entropy. Generally, it corresponds to the bits required to encode the correct answer using the model's distribution. In the case of \relaxed{} we have:
\begin{align*}
    \text{H}(q,P) = -\sum_{i=0}^n log P(A_i \mid B_i)
\end{align*}
This is the cross entropy between two distributions, $q$ and $P$, where $q$ is the delta distribution of the target appearing in the correct position, and $P$ the model's distribution. Thus, this could be understood as the bits required to encode the correct answer \emph{anywhere} in the completion (up to $n$ steps), using the model's (skewed) distribution.

See Table \ref{tab:relaxed_logprobs_example} for an example usage to evaluate model factuality on healthcare benchmarks. Here, we report Relaxed-CrossEntropy instead of \relaxed{}.

%% file: latex/tables/olaph_examples.tex
\begin{table*}[h]
\centering
\renewcommand{\arraystretch}{1.5} 
\resizebox{1.97\columnwidth}{!}{%

\begin{tabular}{@{}rllllcc@{}}
\toprule
\textbf{Question} & \textbf{Must have} & \textbf{Nice to have} & \textbf{Benchmark} & \multicolumn{2}{c}{\textbf{Relaxed-CrossEntropy}} \\ 
\cmidrule(lr){5-6}
 &  &  &  & \textbf{Mistral-7B} & \textbf{JSL-MedLlama-3-8B} \\ \midrule
\begin{tabular}[c]{@{}r@{}}A 50-year-old male presents with a history of recurrent kidney stones\\ and osteopenia. He has been taking high-dose vitamin D supplements\\ due to a previous diagnosis of vitamin D deficiency. Laboratory results\\ reveal hypercalcemia and hypercalciuria. What is the likely diagnosis,\\ and what is the treatment?\end{tabular} 
& Vitamin D toxicity & Stop vitamin D supplementation & Medtext & [2.055, 8.229] & [2.639, 4.142] \\ \midrule
Are benign brain tumors serious? & \begin{tabular}[c]{@{}l@{}}Benign brain tumors are not cancerous\\ and do not spread or invade surrounding\\ tissues.\end{tabular} & \begin{tabular}[c]{@{}l@{}}Benign brain tumors grow slowly\\ and often have clear boundaries.\end{tabular} & OLAPH & [12.825, 15.7796] & [11.208, 16.580] \\ \midrule
\multicolumn{1}{l}{\begin{tabular}[c]{@{}l@{}}We evaluated in the emergency room an adult person who is irritable,\\ yawning, complaining of muscle pain and cramps. They are nauseous\\ and have notable tearing. The pupils are dilated. What is the most\\ probable diagnosis?\end{tabular}} 
& Opioid withdrawal & \begin{tabular}[c]{@{}l@{}}Possibly other substance withdrawal\\ symptoms.\end{tabular} & \careqa{}-Open & [4.2512, 24.7192] & [5.812, 26.883] \\ 
\bottomrule
\end{tabular}

} \caption{Open-ended evaluation using Relaxed Perplexity on samples from MedText, OLAPH, and \careqa{}-Open on Mistral-7B-Instruct-v0.3 (Mistral-7B) and JSL-MedLlama-3-8B-v2.0 (JSL-MedLlama-3-8B). Relaxed-CrossEntropy corresponds to $-\sum_{i=0}^n log P(A_i \mid B_i)$. Lower values indicate the model is more likely to output the correct answer at some time in the completion.}
\label{tab:relaxed_logprobs_example}
\end{table*}

%% file: latex/appendix_results.tex
\section{Evaluation Results}

\input{latex/tables/perplexity_open_ended}
\input{latex/tables/perplexity_clinical_note_taking}
\input{latex/tables/perplexity_recommendations_question_entailment_summarization}

\input{latex/tables/relaxed_perplexity_olaph}

\input{latex/tables/prometheus_new_1}

\input{latex/tables/prometheus_new_2}

\input{latex/tables/open_ended_clinical_note_taking}

\input{latex/tables/open_ended_treatment_recomendation}

\input{latex/tables/medical_factuality}
\input{latex/tables/open_ended_careqa_new}

\input{latex/tables/open_ended_medical_questions}
\input{latex/tables/open_ended_question_entailment}
\input{latex/tables/open_ended_summarization}

\input{latex/tables/close_ended}

\input{}

%% file: latex/tables/perplexity_open_ended.tex
\begin{table*}[t]
\centering
\renewcommand{\arraystretch}{1.5} 
\resizebox{2\columnwidth}{!}{%
\begin{tabular}{l|ccc|ccc|ccc}
\toprule
\textbf{Model} & \multicolumn{9}{c}{\textbf{Open-ended Medical Questions}} \\
 & \multicolumn{3}{c|}{\textbf{\careqa{}-Open}} & \multicolumn{3}{c|}{\textbf{MedDialog Raw}} & \multicolumn{3}{c}{\textbf{MediQA2019}} \\
 & \textbf{Bits per Byte $\downarrow$} & \textbf{Byte Perplexity $\downarrow$} & \textbf{Word Perplexity $\downarrow$} & \textbf{Bits per Byte $\downarrow$} & \textbf{Byte Perplexity $\downarrow$} & \textbf{Word Perplexity $\downarrow$} & \textbf{Bits per Byte $\downarrow$} & \textbf{Byte Perplexity $\downarrow$} & \textbf{Word Perplexity $\downarrow$}\\
\midrule 
BioMistral-MedMNX & 1.302 & 2.465 & 467.349 &  1.043 & 2.060 & 74.760 & 0.416 & 1.335 & 6.044 \\
JSL-MedLlama-3-8B-v2.0 & 1.33 & 2.514 & 534.372 & 1.179 & 2.265 & 131.509 & 0.517 & 1.431 & 9.312 \\
Llama3-Med42-8B & 1.311 & 2.482 & 489.199 & 1.069 & 2.097 & 83.115 & 0.405 & 1.324 & 5.754 \\

Meta-Llama-3.1-70B-Instruct & 1.295 & 2.453 & 452.335 & 0.993 & 1.991 & 60.907 & 0.245 & 1.185 & 2.886 \\
Meta-Llama-3.1-8B-Instruct & 1.346 & 2.543 & 573.723 & 1.060 & 2.085 & 80.124 & 0.430 & 1.347 & 6.407 \\
Mistral-7B-Instruct-v0.3 & 1.442 & 2.717 & 907.864 & 1.073 & 2.104 & 84.603 & 0.420 & 1.338 & 6.145 \\
Mixtral-8x7B-Instruct-v0.1 & 1.453 & 2.738 & 956.752 & 1.028 & 2.039 & 70.258 & 0.300 & 1.232 & 3.662 \\
Phi-3-medium-4k-instruct &  1.255 & 2.387 & 375.453 & 1.068 & 2.097 & 82.957 & 0.410 & 1.329 & 5.884 \\
Phi-3-mini-4k-instruct & 1.342 & 2.535 & 566.127 & 1.082 & 2.117 & 87.936 & 0.444 & 1.360 & 6.796 \\
Qwen2-7B-Instruct & 1.468 & 2.766 & 1024.433 & 1.044 & 2.063 & 75.218 & 0.447 & 1.363 & 6.895 \\
Yi-1.5-34B-Chat & 1.533 & 2.893 & 1392.39 & 1.101 & 2.145 & 95.042 & 0.485 & 1.399 & 8.112 \\
Yi-1.5-9B-Chat & 1.537 & 2.901 & 1416.845 & 1.123 & 2.178 & 104.205 & 0.532 & 1.446 & 9.968 \\

\bottomrule
\end{tabular}%
}
\caption{Perplexity results for Open-ended Medical Questions.}
\end{table*}

%% file: latex/tables/perplexity_clinical_note_taking.tex
\begin{table*}[h]
\centering
\renewcommand{\arraystretch}{1.5} 
\resizebox{2\columnwidth}{!}{%
\begin{tabular}{l|ccc|ccc|ccc}
\toprule
\textbf{Model} & \multicolumn{6}{c}{\textbf{Clinical Note-taking}} & \multicolumn{3}{c}{\textbf{Medical factuality}}\\
 & \multicolumn{3}{c|}{\textbf{ACI Bench}} & \multicolumn{3}{c|}{\textbf{MTS Dialog}} & \multicolumn{3}{c}{\textbf{OLAPH}}\\
 & \textbf{Bits per Byte $\downarrow$} & \textbf{Byte Perplexity $\downarrow$} & \textbf{Word Perplexity $\downarrow$} & \textbf{Bits per Byte $\downarrow$} & \textbf{Byte Perplexity $\downarrow$} & \textbf{Word Perplexity $\downarrow$} & \textbf{Bits per Byte $\downarrow$} & \textbf{Byte Perplexity $\downarrow$} & \textbf{Word Perplexity $\downarrow$} \\
\midrule 
BioMistral-MedMNX & 0.601 & 1.517 & 13.894 & 1.059 & 2.083 & 132.827 & 0.447 & 1.363 & 7.138\\
JSL-MedLlama-3-8B-v2.0 & 0.703 & 1.628 & 21.725 & 1.099 & 2.143 & 160.188 & 0.523& 1.437 & 9.978  \\
Llama3-Med42-8B & 0.485 & 1.399 & 8.357 & 1.060 & 2.085 & 133.416 & 0.450 & 1.366 & 7.211 \\
Meta-Llama-3.1-70B-Instruct & - & - & - & 0.984 & 1.978 & 93.943 & 2.202 & 4.601 & 15946.837 \\
Meta-Llama-3.1-8B-Instruct & 0.612 & 1.529 & 14.618 & 1.074 & 2.105 & 142.211 & 2.181 & 4.533 & 14513.067 \\
Mistral-7B-Instruct-v0.3 & 0.596 & 1.512 & 13.628 & 1.053 & 2.074 & 129.076 & 0.438 & 1.355 & 6.858 \\
Mixtral-8x7B-Instruct-v0.1 & 0.566 & 1.481 & 11.933 & 1.046 & 2.064 & 125.070 & 3.643 & 12.497 & 8992823.856 \\
Phi-3-medium-4k-instruct & 0.642 & 1.560 & 16.600 & 0.971 & 1.960 & 88.447 & 0.393 & 1.313 & 5.620 \\
Phi-3-mini-4k-instruct & 0.599 & 1.514 & 13.754 & 0.972 & 1.962 & 89.163 & 0.407 & 1.326 & 5.986 \\
Qwen2-7B-Instruct & 0.619 & 1.535 & 15.009 & 1.063 & 2.089 & 135.111 & 0.455 & 1.371 & 7.384 \\
Yi-1.5-34B-Chat & 0.728 & 1.657 & 24.270 & 1.099 & 2.143 & 160.265 & 2.798 & 6.955 & 218855.290 \\
Yi-1.5-9B-Chat & 0.711 & 1.636 & 22.456 & 1.180 & 2.265 & 232.073 & 0.571 & 1.485 & 12.281 \\
\bottomrule
\end{tabular}%
}
\caption{Perplexity results for clinical note-taking and medical factuality.}
\end{table*}

%% file: latex/tables/perplexity_recommendations_question_entailment_summarization.tex
\begin{table*}[h]
\centering
\renewcommand{\arraystretch}{1.5} 
\resizebox{2\columnwidth}{!}{%
\begin{tabular}{l|ccc|ccc|ccc}
\toprule
\textbf{Model} & \multicolumn{3}{c|}{\textbf{Making treatment recommendations}} & \multicolumn{3}{c|}{\textbf{Question Entailment}} & \multicolumn{3}{c}{\textbf{Summarization}} \\
 & \multicolumn{3}{c|}{\textbf{MedText}} & \multicolumn{3}{c|}{\textbf{MedDialog Qsumm}} & \multicolumn{3}{c}{\textbf{Mimic-III}} \\
 & \textbf{Bits per Byte $\downarrow$} & \textbf{Byte Perplexity $\downarrow$} & \textbf{Word Perplexity $\downarrow$} & \textbf{Bits per Byte $\downarrow$} & \textbf{Byte Perplexity $\downarrow$} & \textbf{Word Perplexity $\downarrow$} & \textbf{Bits per Byte $\downarrow$} & \textbf{Byte Perplexity $\downarrow$} & \textbf{Word Perplexity $\downarrow$} \\
\midrule 
BioMistral-MedMNX & 0.499 & 1.413 & 10.605 & 1.471 & 2.772 & 275.846 & 1.771 & 3.413 & 4697.580 \\
JSL-MedLlama-3-8B-v2.0 & 0.556 & 1.470 & 13.868 & 1.715 & 3.282 & 699.785 & 2.035 & 4.099 & 16607.943 \\
Llama3-Med42-8B & 0.455 & 1.370 & 8.593 & 1.359 & 2.564 & 179.527 & 1.839 & 3.577 & 6489.224 \\
Meta-Llama-3.1-70B-Instruct & 0.447 & 1.364 & 8.298 & 1.280 & 2.428 & 132.988 &  - & - & - \\
Meta-Llama-3.1-8B-Instruct & 0.534 & 1.448 & 12.501 & 1.371 & 2.587 & 188.513 & 1.826 & 3.545 & 6106.099 \\
Mistral-7B-Instruct-v0.3 & 0.510 & 1.424 & 11.163 & 1.447 & 2.727 & 251.938 & 1.790 & 3.457 & 5138.524 \\
Mixtral-8x7B-Instruct-v0.1 & 0.491 & 1.405 & 10.194 & 1.370 & 2.586 & 187.912 & 1.679 & 3.202 & 3028.534 \\
Phi-3-medium-4k-instruct & 0.423 & 1.341 & 7.400 & 1.332 & 2.517 & 162.163 & 2.084 & 4.239 & 20901.351 \\
Phi-3-mini-4k-instruct & 0.438 & 1.355 & 7.956 & 1.311 & 2.481 & 149.718 & 1.902 & 3.737 & 8784.663 \\
Qwen2-7B-Instruct & 0.527 & 1.441 & 12.106 & 1.383 & 2.608 & 197.167 & 1.878 & 3.676 & 7839.132 \\
Yi-1.5-34B-Chat & 0.556 & 1.470 & 13.875 & 1.437 & 2.708 & 242.427 & 2.202  & 4.600 & 36704.322 \\
Yi-1.5-9B-Chat & 0.559 & 1.473 & 14.052 & 1.470 & 2.771 & 275.222 & 2.341 & 5.067 & 71436.330 \\
\bottomrule
\end{tabular}%
}
\caption{Perplexity results for the following tasks: making diagnosis and treatment recommendation, question entailment and summarization tasks.}
\end{table*}

%% file: latex/tables/relaxed_perplexity_olaph.tex
\begin{table*}[h]
\centering
\renewcommand{\arraystretch}{1.5} 
\resizebox{0.7\columnwidth}{!}{%
\begin{tabular}{l|ccc|ccc}
\toprule
\textbf{Model} & \multicolumn{2}{c}{\textbf{Medical factuality}}\\
&  \multicolumn{2}{c}{\textbf{OLAPH}}\\
 & \textbf{Relaxed perplexity logprobs $\uparrow$} & \textbf{Relaxed perplexity $\downarrow$}\\
\midrule 

BioMistral-MedMNX & -33.122 & 81.532 \\
JSL-MedLlama-3-8B-v2.0 & -39.281 & 12.324 \\
Llama3-Med42-8B & -37.015 & 32.38 \\
Meta-Llama-3.1-70B-Instruct & - & - \\
Meta-Llama-3.1-8B-Instruct & -35.989 & 129.07 \\
Mistral-7B-Instruct-v0.3 & -34.513 & 27.64 \\
Mixtral-8x7B-Instruct-v0.1 & -33.810 & 23.045 \\
Phi-3-medium-4k-instruct & -33.157 & 44.207 \\
Phi-3-mini-4k-instruct & -33.567 & 74.641 \\
Qwen2-7B-Instruct & -37.247 & 133.359 \\
Yi-1.5-34B-Chat & -44.076 & 198.635 \\
Yi-1.5-9B-Chat & -44.501 & 352.381 \\
\bottomrule
\end{tabular}%
}
\caption{Relaxed perplexity results for medical factuality.}
\end{table*}

%% file: latex/tables/prometheus_new_1.tex
\begin{table*}[h]
\centering
\renewcommand{\arraystretch}{1.5} 
\resizebox{2\columnwidth}{!}{%
\begin{tabular}{l|c|c|c|c|c}
\toprule
\textbf{Model} & \textbf{Question Entailment} & \multicolumn{3}{c|}{\textbf{Open-ended Medical Questions}}& \textbf{Treatment recommendations}\\
 & \textbf{MedDialog Qsumm} & \textbf{MedDialog Raw} & \textbf{MediQA2019} & \textbf{\careqa{}-Open}& \textbf{MedText} \\
& \multicolumn{5}{c}{\textbf{Prometheus $\uparrow$}}\\
\midrule 
BioMistral-MedMNX           & 0.163 ± 0.005 & 0.330 ± 0.016 & 0.273 ± 0.027 & 0.240 ± 0.007  & 0.297 ± 0.009  \\
JSL-MedLlama-3-8B-v2.0      & 0.087 ± 0.004 & 0.298 ± 0.017 & 0.365 ± 0.031 & 0.302 ± 0.008 & 0.172 ± 0.008 \\
Llama3-Med42-8B            & 0.241 ± 0.007 & 0.213 ± 0.016 & 0.157 ± 0.024 & 0.105 ± 0.005  & 0.130 ± 0.008 \\
Meta-Llama-3.1-70B-Instruct  & 0.314 ± 0.007 & 0.342 ± 0.016 & 0.313 ± 0.026 & 0.313 ± 0.007  & 0.281 ± 0.009 \\
Meta-Llama-3.1-8B-Instruct   & 0.156 ± 0.005 & 0.263 ± 0.015 & 0.245 ± 0.027 & 0.227 ± 0.007 & 0.237 ± 0.008  \\
Mistral-7B-Instruct-v0.3     & 0.194 ± 0.006 & 0.187 ± 0.015 & 0.087 ± 0.018 & 0.088 ± 0.005 & 0.055 ± 0.005 \\
Mixtral-8x7B-Instruct-v0.1   & 0.112 ± 0.005 & 0.252 ± 0.016 & 0.090 ± 0.017 & 0.130 ± 0.006 & 0.198 ± 0.009 \\
Phi-3-medium-4k-instruct     & 0.168 ± 0.005 & 0.358 ± 0.017 & 0.190 ± 0.023 & 0.319 ± 0.008  & 0.219 ± 0.008  \\
Phi-3-mini-4k-instruct       & 0.126 ± 0.005 & 0.376 ± 0.016 & 0.287 ± 0.027 & 0.185 ± 0.007 & 0.280 ± 0.009 \\
Qwen2-7B-Instruct            & 0.177 ± 0.006 & 0.267 ± 0.014 & 0.255 ± 0.026 & 0.462 ± 0.008 & 0.144 ± 0.007  \\
Yi-1.5-34B-Chat              & 0.179 ± 0.006 & 0.372 ± 0.016 & 0.342 ± 0.030 & 0.492 ± 0.008 & 0.420 ± 0.008  \\
Yi-1.5-9B-Chat               & 0.405 ± 0.007            & 0.550 ± 0.015 & 0.362 ± 0.026 & 0.588 ± 0.007 & 0.397 ± 0.008 \\

\bottomrule
\end{tabular}%
}
\caption{Prometheus results for the following tasks: question entailment, open-ended medical questions and treatment recommendations. }
\end{table*}

%% file: latex/tables/prometheus_new_2.tex
\begin{table*}[h]
\centering
\renewcommand{\arraystretch}{1.7} 
\resizebox{1\columnwidth}{!}{%
\begin{tabular}{l|c|c|c}
\toprule
\textbf{Model} & \textbf{Summarization} & \multicolumn{2}{c}{\textbf{Clinical Note-Taking}}  \\
& \textbf{Mimic-III} & \textbf{MTS Dialog} & \textbf{ACI Bench} \\
& \multicolumn{3}{c}{\textbf{Prometheus $\uparrow$}}\\
\midrule 
BioMistral-MedMNX  &0.535 ± 0.005 & 0.342 ± 0.007 & 0.225 ± 0.063 \\
JSL-MedLlama-3-8B-v2.0   & 0.304 ± 0.005 & 0.459 ± 0.008 & 0.263 ± 0.084\\
Llama3-Med42-8B             & 0.138 ± 0.062 & 0.241 ± 0.007 & 0.138 ± 0.062  \\
Meta-Llama-3.1-70B-Instruct  &  0.293 ± 0.005 & 0.326 ± 0.008 & 0.062 ± 0.043 \\
Meta-Llama-3.1-8B-Instruct   & 0.375 ± 0.005 & 0.229 ± 0.007 & 0.188 ± 0.063\\
Mistral-7B-Instruct-v0.3  & 0.476 ± 0.005 & 0.384 ± 0.008 & 0.050 ± 0.029\\
Mixtral-8x7B-Instruct-v0.1   & 0.543 ± 0.005 & 0.361 ± 0.008 & 0.075 ± 0.036\\
Phi-3-medium-4k-instruct  & 0.249 ± 0.005 & 0.281 ± 0.008 & 0.175 ± 0.064\\
Phi-3-mini-4k-instruct & 0.353 ± 0.005 & 0.328 ± 0.008 & 0.125 ± 0.057\\
Qwen2-7B-Instruct   & 0.541 ± 0.005 & 0.267 ± 0.007 & 0.125 ± 0.052\\
Yi-1.5-34B-Chat & 0.508 ± 0.005 & 0.347 ± 0.009 & 0.287 ± 0.069\\
Yi-1.5-9B-Chat & 0.288 ± 0.005 & 0.417 ± 0.009 & 0.138 ± 0.067\\

\bottomrule
\end{tabular}%
}
\caption{Prometheus results for summarization and clinical-note taking tasks.
}
\end{table*}

%% file: latex/tables/open_ended_clinical_note_taking.tex
\begin{table*}[h]
\centering
\renewcommand{\arraystretch}{1.5} 
\resizebox{2.1\columnwidth}{!}{%
\begin{tabular}{l|ccccccc|ccccccc}
\toprule
\textbf{Model} & \multicolumn{14}{c}{\textbf{Clinical Note-taking}} \\ 
& \multicolumn{7}{c|}{\textbf{ACI Bench}} & \multicolumn{7}{c}{\textbf{MTS Dialog}} \\
 & \textbf{BERTScore $\uparrow$} & \textbf{BLEU $\uparrow$} & \textbf{BLEURT $\uparrow$} & \textbf{MoverScore $\uparrow$} & \textbf{ROUGE1 $\uparrow$} & \textbf{ROUGE2 $\uparrow$} & \textbf{ROUGEL $\uparrow$} & \textbf{BERTScore $\uparrow$} & \textbf{BLEU $\uparrow$} & \textbf{BLEURT $\uparrow$} & \textbf{MoverScore $\uparrow$} & \textbf{ROUGE1 $\uparrow$} & \textbf{ROUGE2 $\uparrow$} & \textbf{ROUGEL $\uparrow$} \\
\midrule
BioMistral-MedMNX & 0.839 ± 0.007 & 0.012 ± 0.005 & -0.834 ± 0.057 & 0.537 ± 0.006 & 0.171 ± 0.016 & 0.039 ± 0.009 & 0.130 ± 0.014 & 0.800 ± 0.001 & 0.001 ± 0.000 & -1.304 ± 0.006 & 0.493 ± 0.001 & 0.040 ± 0.001 & 0.003 ± 0.000 & 0.036 ± 0.001 \\
JSL-MedLlama-3-8B-v2.0 & 0.853 ± 0.011 & 0.033 ± 0.016 & -0.810 ± 0.143 & 0.549 ± 0.013 & 0.212 ± 0.050 & 0.083 ± 0.026 & 0.173 ± 0.040 & 0.801 ± 0.001 & 0.002 ± 0.000 & -1.279 ± 0.007 & 0.492 ± 0.001 & 0.048 ± 0.001 & 0.006 ± 0.001 & 0.043 ± 0.001 \\
Llama3-Med42-8B & 0.863 ± nan & 0.059 ± 0.019 & -0.608 ± nan & 0.564 ± nan & 0.285 ± nan & 0.114 ± nan & 0.224 ± nan & 0.803 ± 0.001 & 0.003 ± 0.001 & -1.290 ± 0.011 & 0.495 ± 0.001 & 0.048 ± 0.002 & 0.007 ± 0.001 & 0.043 ± 0.002 \\
Meta-Llama-3.1-70B-Instruct & 0.852 ± nan & 0.019 ± nan & -0.613 ± nan & 0.548 ± nan & 0.201 ± nan & 0.056 ± nan & 0.154 ± nan & 0.798 ± 0.001 & 0.000 ± 0.000 & -1.350 ± 0.007 & 0.492 ± 0.001 & 0.041 ± 0.001 & 0.002 ± 0.000 & 0.038 ± 0.001 \\
Meta-Llama-3.1-8B-Instruct & 0.829 ± 0.011 & 0.017 ± 0.007 & -0.870 ± 0.068 & 0.538 ± 0.006 & 0.188 ± 0.024 & 0.047 ± 0.013 & 0.138 ± 0.019 & 0.797 ± 0.001 & 0.001 ± 0.000 & -1.364 ± 0.007 & 0.490 ± 0.001 & 0.044 ± 0.001 & 0.003 ± 0.000 & 0.040 ± 0.001 \\
Mistral-7B-Instruct-v0.3 & 0.812 ± nan & 0.000 ± nan & -1.138 ± nan & 0.522 ± nan & 0.046 ± nan & 0.004 ± nan & 0.037 ± nan & 0.800 ± 0.001 & 0.000 ± 0.000 & -1.322 ± 0.007 & 0.491 ± 0.001 & 0.042 ± 0.001 & 0.002 ± 0.000 & 0.039 ± 0.001 \\
Mixtral-8x7B-Instruct-v0.1 & 0.832  ± nan	 &  0.013  ± 0.006	&  -0.881  ± nan & 0.540  ± nan & 0.168  ± nan	& 0.038  ± nan	& 0.119  ± nan & 0.800  ± 0.001	& 0.001  ± 0.000	& -1.349  ± 0.007 & 0.492  ± 0.001
 & 0.042  ± 0.001 & 0.003  ± 0.000	& 0.039  ± 0.001 \\
Phi-3-medium-4k-instruct & 0.824 ± 0.007 & 0.014 ± 0.014 & -1.005 ± 0.067 & 0.528 ± 0.005 & 0.111 ± 0.023 & 0.023 ± 0.011 & 0.086 ± 0.017 & 0.800 ± 0.001 & 0.001 ± 0.000 & -1.346 ± 0.007 & 0.494 ± 0.001 & 0.040 ± 0.001 & 0.003 ± 0.000 & 0.037 ± 0.001 \\
Phi-3-mini-4k-instruct & 0.821 ± nan & 0.015 ± 0.007 & -1.026 ± nan & 0.529 ± nan & 0.135 ± nan & 0.035 ± nan & 0.111 ± nan & 0.800 ± 0.001 & 0.000 ± 0.000 & -1.312 ± 0.007 & 0.494 ± 0.001 & 0.039 ± 0.001 & 0.002 ± 0.000 & 0.036 ± 0.001 \\
Qwen2-7B-Instruct & 0.841 ± nan & 0.015 ± 0.007 & -0.861 ± nan & 0.538 ± nan & 0.167 ± nan & 0.051 ± nan & 0.133 ± nan & 0.798 ± 0.001 & 0.000 ± 0.000 & -1.334 ± 0.006 & 0.489 ± 0.001 & 0.040 ± 0.001 & 0.002 ± 0.000 & 0.037 ± 0.001 \\
Yi-1.5-34B-Chat & 0.840 ± 0.009 & 0.015 ± 0.009 & -0.814 ± 0.085 & 0.533 ± 0.007 & 0.163 ± 0.024 & 0.046 ± 0.015 & 0.126 ± 0.019 & 0.806 ± 0.001 & 0.004 ± 0.001 & -1.266 ± 0.011 & 0.498 ± 0.001 & 0.063 ± 0.003 & 0.012 ± 0.001 & 0.056 ± 0.002 \\
Yi-1.5-9B-Chat & 0.836 ± nan & 0.030 ± 0.024 & -0.892 ± nan & 0.531 ± nan & 0.159 ± nan & 0.063 ± nan & 0.140 ± nan & 0.803 ± 0.001 & 0.003 ± 0.001 & -1.320 ± 0.009 & 0.494 ± 0.001 & 0.053 ± 0.002 & 0.007 ± 0.001 & 0.048 ± 0.002 \\
\bottomrule
\end{tabular}%
}
\caption{Clinical note-taking results.}
\label{tab:evaluation_results}
\end{table*}

%% file: latex/tables/open_ended_treatment_recomendation.tex
\begin{table*}[h]
\centering
\renewcommand{\arraystretch}{1.5} 
\resizebox{2.1\columnwidth}{!}{%
\begin{tabular}{l|ccccccc}
\toprule
\multirow{2}{*}{\textbf{Model}} & \multicolumn{7}{c}{\textbf{Making Treatment Recommendations}} \\  
                                  & \multicolumn{7}{c}{\textbf{Medtext}} \\ 
                                  & \textbf{BERTScore $\uparrow$} & \textbf{BLEU $\uparrow$} & \textbf{BLEURT $\uparrow$} & \textbf{MoverScore $\uparrow$} & \textbf{ROUGE1 $\uparrow$} & \textbf{ROUGE2 $\uparrow$} & \textbf{ROUGEL $\uparrow$} \\ 
\midrule
BioMistral-MedMNX & 0.855 ± 0.001 & 0.013 ± 0.001 & -0.650 ± 0.007 & 0.547 ± 0.001 & 0.177 ± 0.002 & 0.037 ± 0.001 & 0.136 ± 0.002 \\
JSL-MedLlama-3-8B-v2.0 & 0.856 ± 0.001 & 0.021 ± 0.002 & -0.652 ± 0.012 & 0.546 ± 0.001 & 0.185 ± 0.003 & 0.045 ± 0.002 & 0.146 ± 0.003 \\
Llama3-Med42-8B & 0.865 ± 0.001 & 0.018 ± 0.002 & -0.546 ± 0.015 & 0.557 ± 0.001 & 0.204 ± 0.005 & 0.052 ± 0.003 & 0.158 ± 0.004 \\
Meta-Llama-3.1-70B-Instruct & 0.859 ± 0.001 & 0.022 ± 0.002 & -0.644 ± 0.008 & 0.547 ± 0.001 & 0.196 ± 0.003 & 0.048 ± 0.002 & 0.150 ± 0.002 \\
Meta-Llama-3.1-8B-Instruct & 0.843 ± 0.001 & 0.010 ± 0.001 & -0.839 ± 0.007 & 0.535 ± 0.001 & 0.155 ± 0.002 & 0.032 ± 0.001 & 0.120 ± 0.002 \\
Mistral-7B-Instruct-v0.3 & 0.870 ± 0.002 & 0.038 ± 0.005 & -0.467 ± 0.022 & 0.562 ± 0.002 & 0.230 ± 0.008 & 0.072 ± 0.006 & 0.183 ± 0.007 \\
Mixtral-8x7B-Instruct-v0.1 & 0.868  ± 0.001	& 0.029  ± 0.002	& -0.502  ± 0.011	& 0.559  ± 0.001	& 0.220  ± 0.003	& 0.060  ± 0.002	& 0.172  ± 0.003 \\ 
Phi-3-medium-4k-instruct & 0.869 ± 0.001 & 0.033 ± 0.002 & -0.504 ± 0.011 & 0.560 ± 0.001 & 0.231 ± 0.004 & 0.069 ± 0.003 & 0.182 ± 0.003 \\
Phi-3-mini-4k-instruct & 0.863 ± 0.001 & 0.027 ± 0.002 & -0.551 ± 0.009 & 0.555 ± 0.001 & 0.213 ± 0.003 & 0.060 ± 0.002 & 0.165 ± 0.003 \\
Qwen2-7B-Instruct & 0.859 ± 0.001 & 0.021 ± 0.002 & -0.634 ± 0.012 & 0.547 ± 0.001 & 0.193 ± 0.004 & 0.049 ± 0.002 & 0.147 ± 0.003 \\
Yi-1.5-34B-Chat & 0.867 ± 0.001	& 0.033 ± 0.002	& -0.580 ± 0.008	& 0.559 ± 0.001	& 0.245 ± 0.003	& 0.074 ± 0.002	& 0.189 ± 0.002 \\
Yi-1.5-9B-Chat & 0.863 ± 0.000 & 0.022 ± 0.001 & -0.513 ± 0.006 & 0.555 ± 0.001 & 0.213 ± 0.002 & 0.054 ± 0.002 & 0.163 ± 0.002 \\
\bottomrule
\end{tabular}}
\caption{Making diagnosis and treatment recommendations results.}
\label{tab:treatment_recommendations}
\end{table*}

%% file: latex/tables/medical_factuality.tex
\begin{table*}[h]
\centering
\renewcommand{\arraystretch}{1.5} 
\resizebox{2.1\columnwidth}{!}{%
\begin{tabular}{l|ccccccc}
\toprule
\multirow{2}{*}{\textbf{Model}} & \multicolumn{7}{c}{\textbf{Medical factuality}} \\  
                                  & \multicolumn{7}{c}{\textbf{OLAPH}} \\ 
                                  & \textbf{BERTScore $\uparrow$} & \textbf{BLEU $\uparrow$} & \textbf{BLEURT $\uparrow$} & \textbf{MoverScore $\uparrow$} & \textbf{ROUGE1 $\uparrow$} & \textbf{ROUGE2 $\uparrow$} & \textbf{ROUGEL $\uparrow$} \\ 
\midrule
BioMistral-MedMNX & 0.864 ± 0.001 & 0.022 ± 0.002 & -0.557 ± 0.014 & 0.555 ± 0.001 & 0.211 ± 0.004 & 0.058 ± 0.002 & 0.166 ± 0.003 \\
JSL-MedLlama-3-8B-v2.0 & 0.868 ± 0.001 & 0.031 ± 0.003 & -0.544 ± 0.019 & 0.558 ± 0.002 & 0.230 ± 0.005 & 0.071 ± 0.004 & 0.183 ± 0.005 \\
Llama3-Med42-8B & 0.876 ± 0.001 & 0.024 ± 0.002 & -0.387 ± 0.015 & 0.567 ± 0.001 & 0.239 ± 0.005 & 0.069 ± 0.004 & 0.185 ± 0.005 \\
Meta-Llama-3.1-70B-Instruct & 0.866 ± 0.001 & 0.021 ± 0.002 & -0.538 ± 0.017 & 0.559 ± 0.001 & 0.225 ± 0.005 & 0.064 ± 0.004 & 0.178 ± 0.005 \\
Meta-Llama-3.1-8B-Instruct & 0.845 ± 0.001 & 0.009 ± 0.001 & -0.792 ± 0.015 & 0.538 ± 0.001 & 0.166 ± 0.004 & 0.038 ± 0.002 & 0.129 ± 0.003 \\
Mistral-7B-Instruct-v0.3 & 0.886 ± 0.001 & 0.056 ± 0.005 & -0.285 ± 0.022 & 0.581 ± 0.002 & 0.293 ± 0.008 & 0.110 ± 0.006 & 0.240 ± 0.007 \\
Mixtral-8x7B-Instruct-v0.1 & 0.810 ± 0.003 & 0.000 ± 0.000 & -1.148 ± 0.015 & 0.501 ± 0.001 & 0.081 ± 0.004 & 0.003 ± 0.001 & 0.067 ± 0.003 \\
Phi-3-medium-4k-instruct & 0.880 ± 0.002 & 0.047 ± 0.005 & -0.369 ± 0.022 & 0.574 ± 0.002 & 0.274 ± 0.007 & 0.096 ± 0.006 & 0.221 ± 0.007 \\
Phi-3-mini-4k-instruct & 0.867 ± 0.002 & 0.025 ± 0.003 & -0.494 ± 0.022 & 0.559 ± 0.002 & 0.220 ± 0.007 & 0.063 ± 0.004 & 0.177 ± 0.006 \\
Qwen2-7B-Instruct & 0.876 ± 0.001 & 0.033 ± 0.003 & -0.349 ± 0.014 & 0.570 ± 0.001 & 0.250 ± 0.005 & 0.076 ± 0.003 & 0.200 ± 0.004 \\
Yi-1.5-34B-Chat & 0.879 ± 0.001 & 0.041 ± 0.003 & -0.371 ± 0.016 & 0.570 ± 0.002 & 0.269 ± 0.006 & 0.092 ± 0.004 & 0.216 ± 0.005 \\
Yi-1.5-9B-Chat & 0.878 ± 0.001 & 0.037 ± 0.002 & -0.349 ± 0.012 & 0.569 ± 0.001 & 0.253 ± 0.004 & 0.083 ± 0.003 & 0.203 ± 0.004 \\
\bottomrule
\end{tabular}}
\caption{Medical factuality results.}
\label{tab:medical_factuality}
\end{table*}

%% file: latex/tables/open_ended_careqa_new.tex
\begin{table*}[h]
\centering
\renewcommand{\arraystretch}{1.5} 
\resizebox{2.1\columnwidth}{!}{%
\begin{tabular}{l|ccccccc}
\toprule
\textbf{Model} & \multicolumn{7}{c}{\textbf{Open-ended medical questions}} \\ 
& \multicolumn{7}{c}{\textbf{\careqa{}-Open}} \\
& \textbf{BERTScore $\uparrow$} & \textbf{BLEU $\uparrow$} & \textbf{BLEURT $\uparrow$} & \textbf{MoverScore $\uparrow$} & \textbf{ROUGE1 $\uparrow$} & \textbf{ROUGE2 $\uparrow$} & \textbf{ROUGEL $\uparrow$} \\  
\midrule
    BioMistral-MedMNX & 0.816 ± 0.002 & 0.002 ± 0.000 & -1.329 ± 0.009 & 0.492 ± 0.001 & 0.066 ± 0.002 & 0.017 ± 0.001 & 0.058 ± 0.002 \\ 
    JSL-MedLlama-3-8B-v2.0 & 0.827 ± 0.001 & 0.003 ± 0.000 & -1.234 ± 0.009 & 0.493 ± 0.001 & 0.069 ± 0.002 & 0.019 ± 0.001 & 0.060 ± 0.002 \\ 
    Llama3-Med42-8B & 0.293 ± 0.010 & 0.002 ± 0.001 & -1.441 ± 0.010 & 0.503 ± 0.001 & 0.030 ± 0.002 & 0.006 ± 0.001 & 0.027 ± 0.002 \\ 
    Meta-Llama-3.1-70B-Instruct & 0.660 ± 0.007 & 0.005 ± 0.001 & -1.283 ± 0.010 & 0.508 ± 0.001 & 0.096 ± 0.003 & 0.031 ± 0.002 & 0.087 ± 0.003 \\ 
    Meta-Llama-3.1-8B-Instruct & 0.761 ± 0.004 & 0.002 ± 0.000 & -1.496 ± 0.007 & 0.485 ± 0.001 & 0.049 ± 0.001 & 0.013 ± 0.001 & 0.042 ± 0.001 \\ 
    Mistral-7B-Instruct-v0.3 & 0.841 ± 0.002 & 0.004 ± 0.001 & -1.212 ± 0.026 & 0.501 ± 0.003 & 0.109 ± 0.008 & 0.037 ± 0.006 & 0.098 ± 0.008 \\ 
    Mixtral-8x7B-Instruct-v0.1 & 0.768 ± 0.010 & 0.008 ± 0.001 & -1.140 ± 0.022 & 0.515 ± 0.003 & 0.126 ± 0.007 & 0.040 ± 0.004 & 0.114 ± 0.007 \\ 
    Phi-3-medium-4k-instruct & 0.814 ± 0.003 & 0.005 ± 0.001 & -1.276 ± 0.010 & 0.499 ± 0.001 & 0.089 ± 0.003 & 0.028 ± 0.001 & 0.077 ± 0.002 \\ 
    Phi-3-mini-4k-instruct & 0.684 ± 0.008 & 0.003 ± 0.001 & -1.277 ± 0.010 & 0.500 ± 0.001 & 0.064 ± 0.002 & 0.016 ± 0.001 & 0.054 ± 0.002 \\ 
    Qwen2-7B-Instruct & 0.755 ± 0.005 & 0.003 ± 0.000 & -1.229 ± 0.008 & 0.496 ± 0.001 & 0.067 ± 0.002 & 0.018 ± 0.001 & 0.057 ± 0.001 \\ 
    Yi-1.5-34B-Chat & 0.809 ± 0.003 & 0.005 ± 0.001 & -1.186 ± 0.008 & 0.496 ± 0.001 & 0.078 ± 0.002 & 0.024 ± 0.001 & 0.067 ± 0.002 \\ 
    Yi-1.5-9B-Chat & 0.831 ± 0.001 & 0.004 ± 0.000 & -1.180 ± 0.008 & 0.491 ± 0.001 & 0.079 ± 0.002 & 0.023 ± 0.001 & 0.066 ± 0.002 \\ 
\bottomrule
\end{tabular}
} 
\caption{Results for \careqa{}-Open.}
\label{tab:question_entailment_results}
\end{table*}

%% file: latex/tables/open_ended_medical_questions.tex
\begin{table*}[h]
\centering
\renewcommand{\arraystretch}{1.5} 
\resizebox{2.1\columnwidth}{!}{%

\begin{tabular}{l|ccccccc|ccccccc}
\toprule
\textbf{Model} & \multicolumn{14}{c}{\textbf{Open-ended Medical Questions}} \\ 
& \multicolumn{7}{c|}{\textbf{MedDialog Raw}} & \multicolumn{7}{c}{\textbf{MEDIQA2019}} \\
& \textbf{BERTScore $\uparrow$} & \textbf{BLEU $\uparrow$} & \textbf{BLEURT $\uparrow$} & \textbf{MoverScore $\uparrow$} & \textbf{ROUGE1 $\uparrow$} & \textbf{ROUGE2 $\uparrow$} & \textbf{ROUGEL $\uparrow$} & \textbf{BERTScore $\uparrow$} & \textbf{BLEU $\uparrow$} & \textbf{BLEURT $\uparrow$} & \textbf{MoverScore $\uparrow$} & \textbf{ROUGE1 $\uparrow$} & \textbf{ROUGE2 $\uparrow$} & \textbf{ROUGEL $\uparrow$} \\ 
\midrule
BioMistral-MedMNX & 0.833 ± 0.001 & 0.001 ± 0.000 & -0.898 ± 0.012 & 0.526 ± 0.001 & 0.113 ± 0.003 & 0.010 ± 0.001 & 0.088 ± 0.002 & 0.850 ± 0.002 & 0.005 ± 0.002 & -0.660 ± 0.024 & 0.547 ± 0.002 & 0.169 ± 0.007 & 0.032 ± 0.003 & 0.132 ± 0.005 \\
JSL-MedLlama-3-8B-v2.0 & 0.832 ± 0.001 & 0.000 ± 0.000 & -0.875 ± 0.015 & 0.524 ± 0.001 & 0.109 ± 0.003 & 0.009 ± 0.001 & 0.087 ± 0.002 & 0.849 ± 0.002 & 0.008 ± 0.002 & -0.688 ± 0.027 & 0.543 ± 0.002 & 0.164 ± 0.006 & 0.030 ± 0.003 & 0.130 ± 0.005 \\
Llama3-Med42-8B & 0.834 ± 0.001 & 0.000 ± 0.000 & -0.887 ± 0.019 & 0.527 ± 0.001 & 0.108 ± 0.004 & 0.010 ± 0.001 & 0.085 ± 0.003 & 0.850 ± 0.003 & 0.008 ± 0.003 & -0.646 ± 0.043 & 0.546 ± 0.004 & 0.166 ± 0.012 & 0.026 ± 0.005 & 0.129 ± 0.010 \\
Meta-Llama-3.1-70B-Instruct & 0.835 ± 0.001 & 0.000 ± 0.000 & -0.875 ± 0.014 & 0.525 ± 0.001 & 0.115 ± 0.003 & 0.011 ± 0.001 & 0.089 ± 0.002 & 0.856 ± 0.002 & 0.010 ± 0.003 & -0.630 ± 0.030 & 0.547 ± 0.002 & 0.176 ± 0.008 & 0.037 ± 0.004 & 0.139 ± 0.007 \\
Meta-Llama-3.1-8B-Instruct & 0.824 ± 0.001 & 0.000 ± 0.000 & -1.013 ± 0.011 & 0.521 ± 0.001 & 0.096 ± 0.003 & 0.008 ± 0.001 & 0.074 ± 0.002 & 0.843 ± 0.002 & 0.005 ± 0.001 & -0.775 ± 0.024 & 0.538 ± 0.002 & 0.154 ± 0.007 & 0.028 ± 0.003 & 0.117 ± 0.005 \\
Mistral-7B-Instruct-v0.3 & 0.841 ± 0.001 & 0.000 ± 0.000 & -0.762 ± 0.024 & 0.530 ± 0.001 & 0.121 ± 0.005 & 0.014 ± 0.002 & 0.095 ± 0.004 & 0.852 ± 0.004 & 0.016 ± 0.008 & -0.661 ± 0.061 & 0.541 ± 0.005 & 0.158 ± 0.016 & 0.046 ± 0.011 & 0.132 ± 0.015 \\

Mixtral-8x7B-Instruct-v0.1 & 0.838  ± 0.001	& 0.001  ± 0.000	& -0.819  ± 0.020	& 0.529  ± 0.001	& 0.119  ± 0.004	& 0.012  ± 0.001	& 0.093  ± 0.003 & 0.846 ± 0.004	& 0.006 ± 0.003	& -0.837 ± 0.058 &	0.536 ± 0.004	& 0.135 ± 0.015	& 0.022 ± 0.009	& 0.110 ± 0.014 \\

Phi-3-medium-4k-instruct & 0.837 ± 0.001 & 0.001 ± 0.000 & -0.854 ± 0.016 & 0.528 ± 0.001 & 0.121 ± 0.004 & 0.013 ± 0.001 & 0.093 ± 0.003 & 0.859 ± 0.003 & 0.011 ± 0.004 & -0.552 ± 0.042 & 0.551 ± 0.004 & 0.197 ± 0.013 & 0.049 ± 0.009 & 0.157 ± 0.012 \\
Phi-3-mini-4k-instruct & 0.834 ± 0.001 & 0.000 ± 0.000 & -0.891 ± 0.013 & 0.526 ± 0.001 & 0.103 ± 0.003 & 0.009 ± 0.001 & 0.082 ± 0.002 & 0.850 ± 0.003 & 0.008 ± 0.004 & -0.682 ± 0.036 & 0.543 ± 0.003 & 0.163 ± 0.011 & 0.032 ± 0.007 & 0.129 ± 0.009 \\
Qwen2-7B-Instruct & 0.833 ± 0.001 & 0.000 ± 0.000 & -0.939 ± 0.015 & 0.526 ± 0.001 & 0.109 ± 0.004 & 0.010 ± 0.001 & 0.084 ± 0.003 & 0.851 ± 0.002 & 0.005 ± 0.002 & -0.673 ± 0.031 & 0.542 ± 0.002 & 0.155 ± 0.008 & 0.029 ± 0.005 & 0.120 ± 0.007 \\
Yi-1.5-34B-Chat & 0.839 ± 0.001 & 0.000 ± 0.000 & -0.785 ± 0.014 & 0.529 ± 0.001 & 0.131 ± 0.004 & 0.016 ± 0.001 & 0.101 ± 0.003 & 0.858 ± 0.002 & 0.008 ± 0.002 & -0.524 ± 0.031 & 0.551 ± 0.003 & 0.185 ± 0.009 & 0.039 ± 0.005 & 0.147 ± 0.008 \\
Yi-1.5-9B-Chat & 0.837 ± 0.001 & 0.001 ± 0.000 & -0.804 ± 0.012 & 0.528 ± 0.001 & 0.123 ± 0.003 & 0.014 ± 0.001 & 0.096 ± 0.002 & 0.857 ± 0.002 & 0.011 ± 0.003 & -0.540 ± 0.026 & 0.549 ± 0.002 & 0.197 ± 0.007 & 0.043 ± 0.004 & 0.159 ± 0.006 \\
\bottomrule
\end{tabular}
} 
\caption{Open-ended medical questions results.}
\label{tab:open_ended_medical_questions}
\end{table*}

%% file: latex/tables/open_ended_question_entailment.tex
\begin{table*}[h]
\centering
\renewcommand{\arraystretch}{1.5} 
\resizebox{2.1\columnwidth}{!}{%
\begin{tabular}{l|ccccccc}
\toprule
\textbf{Model} & \multicolumn{7}{c}{\textbf{Question Entailment}} \\ 
& \multicolumn{7}{c}{\textbf{MedDialog Qsumm}} \\
& \textbf{BERTScore $\uparrow$} & \textbf{BLEU $\uparrow$} & \textbf{BLEURT $\uparrow$} & \textbf{MoverScore $\uparrow$} & \textbf{ROUGE1 $\uparrow$} & \textbf{ROUGE2 $\uparrow$} & \textbf{ROUGEL $\uparrow$} \\  
\midrule
BioMistral-MedMNX & 0.839 ± 0.000 & 0.005 ± 0.000 & -1.056 ± 0.003 & 0.520 ± 0.000 & 0.093 ± 0.001 & 0.018 ± 0.001 & 0.081 ± 0.001 \\
JSL-MedLlama-3-8B-v2.0 & 0.840 ± 0.000 & 0.004 ± 0.000 & -0.967 ± 0.004 & 0.522 ± 0.000 & 0.085 ± 0.001 & 0.013 ± 0.001 & 0.074 ± 0.001 \\
Llama3-Med42-8B & 0.845 ± 0.000 & 0.004 ± 0.000 & -1.020 ± 0.005 & 0.521 ± 0.000 & 0.099 ± 0.002 & 0.019 ± 0.001 & 0.084 ± 0.001 \\
Meta-Llama-3.1-70B-Instruct & 0.849 ± 0.000 & 0.008 ± 0.001 & -1.013 ± 0.005 & 0.525 ± 0.000 & 0.120 ± 0.002 & 0.029 ± 0.001 & 0.102 ± 0.001 \\
Meta-Llama-3.1-8B-Instruct & 0.836 ± 0.000 & 0.005 ± 0.000 & -1.097 ± 0.004 & 0.518 ± 0.000 & 0.091 ± 0.001 & 0.017 ± 0.001 & 0.078 ± 0.001 \\
Mistral-7B-Instruct-v0.3 & 0.852 ± 0.001 & 0.010 ± 0.001 & -0.966 ± 0.007 & 0.526 ± 0.001 & 0.122 ± 0.003 & 0.031 ± 0.002 & 0.106 ± 0.002 \\
Mixtral-8x7B-Instruct-v0.1 & 0.848 ± 0.001	& 0.004 ± 0.000	& -0.984 ± 0.006	& 0.525 ± 0.000& 	0.099 ± 0.002	& 0.020 ± 0.001	& 0.086 ± 0.002 \\
Phi-3-medium-4k-instruct & 0.839 ± 0.000 & 0.004 ± 0.000 & -1.086 ± 0.004 & 0.522 ± 0.000 & 0.093 ± 0.001 & 0.017 ± 0.001 & 0.081 ± 0.001 \\
Phi-3-mini-4k-instruct & 0.840 ± 0.000 & 0.003 ± 0.000 & -1.041 ± 0.004 & 0.521 ± 0.000 & 0.083 ± 0.001 & 0.012 ± 0.001 & 0.072 ± 0.001 \\
Qwen2-7B-Instruct & 0.844 ± 0.000 & 0.006 ± 0.001 & -1.007 ± 0.004 & 0.524 ± 0.000 & 0.102 ± 0.002 & 0.020 ± 0.001 & 0.088 ± 0.001 \\
Yi-1.5-34B-Chat & 0.842 ± 0.001	&0.006 ± 0.001	&-1.010 ± 0.005	&0.522 ± 0.000	&0.100 ± 0.002	&0.021 ± 0.001	&0.087 ± 0.002 \\
Yi-1.5-9B-Chat & 0.852 ± 0.000 & 0.010 ± 0.001 & -0.979 ± 0.004 & 0.525 ± 0.000 & 0.128 ± 0.001 & 0.033 ± 0.001 & 0.109 ± 0.001 \\
\bottomrule
\end{tabular}
} 
\caption{Question entailment results.}
\label{tab:question_entailment_results}
\end{table*}

%% file: latex/tables/open_ended_summarization.tex
\begin{table*}[h]
\centering
\renewcommand{\arraystretch}{1.5} 
\resizebox{2.1\columnwidth}{!}{%
\begin{tabular}{l|cccccccc}
\toprule
\textbf{Model} & \multicolumn{8}{c}{\textbf{Summarization}} \\ 
& \multicolumn{8}{c}{\textbf{MIMIC-III}} \\
&\textbf{F1-RadGraph $\uparrow$} & \textbf{BERTScore $\uparrow$} & \textbf{BLEU $\uparrow$} & \textbf{BLEURT $\uparrow$} & \textbf{MoverScore $\uparrow$} & \textbf{ROUGE1 $\uparrow$} & \textbf{ROUGE2 $\uparrow$} & \textbf{ROUGEL $\uparrow$} \\ 
\midrule
BioMistral-MedMNX & 0.089 ± 0.001 & 0.837 ± 0.000 & 0.009 ± 0.000 & -0.796 ± 0.003 & 0.551 ± 0.000 & 0.130 ± 0.001 & 0.031 ± 0.001 & 0.110 ± 0.001 \\
JSL-MedLlama-3-8B-v2.0 & 0.079 ± 0.002 & 0.841 ± 0.000 & 0.014 ± 0.001 & -0.780 ± 0.005 & 0.556 ± 0.001 & 0.143 ± 0.002 & 0.041 ± 0.001 & 0.124 ± 0.002 \\
Llama3-Med42-8B & 0.093 ± 0.002 & 0.843 ± 0.000 & 0.013 ± 0.001 & -0.729 ± 0.005 & 0.557 ± 0.001 & 0.152 ± 0.002 & 0.041 ± 0.001 & 0.129 ± 0.002 \\
Meta-Llama-3.1-70B-Instruct & 0.059 ± 0.002 & 0.836 ± 0.000 & 0.009 ± 0.001 & -0.811 ± 0.005 & 0.547 ± 0.001 & 0.130 ± 0.002 & 0.031 ± 0.001 & 0.110 ± 0.002 \\
Meta-Llama-3.1-8B-Instruct & 0.065 ± 0.001 & 0.830 ± 0.000 & 0.007 ± 0.000 & -0.834 ± 0.004 & 0.542 ± 0.000 & 0.115 ± 0.001 & 0.025 ± 0.001 & 0.097 ± 0.001 \\
Mistral-7B-Instruct-v0.3 & 0.082 ± 0.002 & 0.845 ± 0.000 & 0.013 ± 0.001 & -0.753 ± 0.005 & 0.558 ± 0.000 & 0.157 ± 0.002 & 0.044 ± 0.001 & 0.134 ± 0.002 \\
Mixtral-8x7B-Instruct-v0.1 &
0.088 ± 0.002	& 0.844 ± 0.000	& 0.015 ± 0.001	& -0.762 ± 0.004	& 0.557 ± 0.000	& 0.157 ± 0.002	& 0.044 ± 0.001	& 0.134 ± 0.002 \\
Phi-3-medium-4k-instruct & 0.038 ± 0.002	& 0.838 ± 0.001	& 0.010 ± 0.001	& -0.771 ± 0.008 & 0.550 ± 0.001	& 0.137 ± 0.003	& 0.034 ± 0.001	& 0.116 ± 0.002 \\ 
Phi-3-mini-4k-instruct & 0.066 ± 0.002	&0.836 ± 0.000	& 0.008 ± 0.001	& -0.767 ± 0.005 & 0.548 ± 0.001	& 0.123 ± 0.002	& 0.029 ± 0.001	& 0.104 ± 0.002 \\ 
Qwen2-7B-Instruct & 0.078 ± 0.001 & 0.843 ± 0.000 & 0.009 ± 0.000 & -0.761 ± 0.004 & 0.555 ± 0.000 & 0.142 ± 0.002 & 0.035 ± 0.001 & 0.120 ± 0.001 \\
Yi-1.5-34B-Chat & 0.065 ± 0.001	& 0.839 ± 0.000	& 0.009 ± 0.001	& -0.775 ± 0.004	& 0.550 ± 0.000 &	0.137 ± 0.002	& 0.033 ± 0.001	& 0.116 ± 0.001 \\ 
Yi-1.5-9B-Chat & 0.080 ± 0.002 & 0.840 ± 0.000 & 0.012 ± 0.001 & -0.806 ± 0.005 & 0.554 ± 0.001 & 0.136 ± 0.002 & 0.035 ± 0.001 & 0.117 ± 0.002 \\
\bottomrule
\end{tabular}
}
\caption{Summarization results.}
\label{tab:summarization_results}
\end{table*}

%% file: latex/tables/close_ended.tex
\begin{table*}[h]
\centering
\renewcommand{\arraystretch}{1.5} 
\resizebox{2.1\columnwidth}{!}{%
\begin{tabular}{l|cccccccccc}
\toprule
\textbf{Model} & \multicolumn{10}{c}{\textbf{Close-ended}} \\ 
 & \textbf{MedMCQA $\uparrow$} & \textbf{MedQA $\uparrow$} & \textbf{\careqa{} (en) $\uparrow$} & \textbf{\careqa{} (es) $\uparrow$} & \textbf{multimedqa $\uparrow$} & \textbf{PubMedQA $\uparrow$} & \textbf{Med Text Classification $\uparrow$} & \textbf{Med Transcriptions $\uparrow$} & \textbf{BioRED $\uparrow$} & \textbf{MMLU $\uparrow$} \\ 
\midrule
BioMistral-MedMNX & 0.495 ± 0.008 & 0.515 ± 0.014 & 0.629 ± 0.006	& 0.546 ± 0.007 & 0.547 ± 0.006 & 0.776 ± 0.019 & 0.202 ± 0.011 & 0.356 ± 0.007 & 0.216 ± 0.013 & 0.6784 ± 0.034 \\
JSL-MedLlama-3-8B-v2.0 & 0.613 ± 0.008 & 0.617 ± 0.014 & 0.672 ± 0.006	& 0.572 ± 0.007  & 0.648 ± 0.006 & 0.742 ± 0.020 & 0.191 ± 0.010 & 0.361 ± 0.007 & 0.254 ± 0.014 & 0.7739 ± 0.0305 \\
Llama3-Med42-8B & 0.603 ± 0.008 & 0.626 ± 0.014 & 0.683 ± 0.006	& 0.575 ± 0.007  & 0.642 ± 0.006 & 0.772 ± 0.019 & 0.202 ± 0.011 & 0.377 ± 0.007 & 0.203 ± 0.013 & 0.7525 ± 0.0315 \\
Meta-Llama-3.1-70B-Instruct & 0.722 ± 0.007 & 0.798 ± 0.011 & 0.837 ± 0.005	& 0.825 ± 0.005 &  0.764 ± 0.005 & 0.800 ± 0.018 & 0.145 ± 0.003 & 0.381 ± 0.007 & 0.515 ± 0.016 & 0.8711 ± 0.0236 \\
Meta-Llama-3.1-8B-Instruct & 0.593 ± 0.008 & 0.637 ± 0.013 & 0.700 ± 0.006	& 0.592 ± 0.007  & 0.638 ± 0.006 & 0.752 ± 0.019 & 0.161 ± 0.003 & 0.334 ± 0.007  & 0.232 ± 0.013 & 0.7621 ± 0.031 \\
Mistral-7B-Instruct-v0.3 & 0.482 ± 0.008 & 0.523 ± 0.014 & 0.607 ± 0.007	& 0.529 ± 0.007   & 0.538 ± 0.006 & 0.774 ± 0.019 & 0.178 ± 0.010 & 0.356 ± 0.007 & 0.358 ± 0.015 & 0.661 ± 0.0345 \\
Mixtral-8x7B-Instruct-v0.1 & 0.564  ± 0.008	&  0.614  ± 0.014 & 0.725 ± 0.006	& 0.688 ± 0.006   & 0.622  ± 0.006 & 0.796  ± 0.018 & 0.207  ± 0.011 & 0.344 ± 0.007 & 0.352  ± 0.015 & 0.7766 ± 0.0304  \\
Phi-3-medium-4k-instruct & 0.623 ± 0.007 & 0.596 ± 0.014 & 0.769 ± 0.006	& 0.718 ± 0.006  & 0.661 ± 0.006 & 0.782 ± 0.018 & 0.048 ± 0.002 & 0.365 ± 0.007 & 0.261 ± 0.014 & 0.8237 ± 0.0275 \\
Phi-3-mini-4k-instruct & 0.572 ± 0.008 & 0.537 ± 0.014 &  0.701 ± 0.006	& 0.585 ± 0.007  & 0.604 ± 0.006 & 0.752 ± 0.019 & 0.192 ± 0.003 & 0.367 ± 0.007 & 0.262 ± 0.014 & 0.7398 ± 0.0321 \\
Qwen2-7B-Instruct & 0.551 ± 0.008 & 0.570 ± 0.014 & 0.680 ± 0.006	& 0.621 ± 0.006   & 0.596 ± 0.006 & 0.742 ± 0.020 & 0.225 ± 0.011 & 0.363 ± 0.007 & 0.197 ± 0.013 & 0.7337 ± 0.032 \\
Yi-1.5-34B-Chat & 0.575 ± 0.008 & 0.614 ± 0.014 & 0.733 ± 0.006	& 0.632 ± 0.006  & 0.628 ± 0.006 & 0.774 ± 0.019 & 0.301 ± 0.012 & 0.345 ± 0.007 & 0.543 ± 0.016 & 0.7806 ± 0.0298 \\
Yi-1.5-9B-Chat & 0.488 ± 0.008 & 0.515 ± 0.014 & 0.650 ± 0.006	& 0.507 ± 0.007  & 0.546 ± 0.006 & 0.774 ± 0.019 & 0.227 ± 0.011 & 0.330 ± 0.007 & 0.537 ± 0.016 & 0.7007 ± 0.0329 \\
\bottomrule
\end{tabular}
} 
\caption{Close-ended results.}
\label{tab:close_ended_benchmarks}
\end{table*}